\pdfoutput=1

\documentclass[12pt]{article}
\usepackage{graphicx,psfrag,epsf}
\usepackage{enumerate}
\usepackage{natbib}
\usepackage{amsthm, amsmath, amsfonts, mathrsfs, amssymb, bm, latexsym}
\usepackage{mathtools}
\usepackage{graphicx}
\usepackage[T1]{fontenc}
\usepackage{caption}
\usepackage{subcaption}
\usepackage{algorithm}
\usepackage{algpseudocode}
\usepackage{float}
\usepackage{times}
\usepackage{xcolor}
\usepackage{multirow}
\usepackage{authblk}
\usepackage{url}
\usepackage{dsfont}

\newcommand{\blind}{0}

\begin{document}

\def\spacingset#1{\renewcommand{\baselinestretch}
{#1}\small\normalsize} \spacingset{1}

\if0\blind
{
  \title{\bf Positional Encoder Graph Quantile Neural Networks for Geographic Data}
  \author[1]{William E.~R.~de Amorim}
  \author[2]{Scott A.~Sisson}
  \author[1]{T. Rodrigues}
  \author[3]{David J.~Nott}
  \author[1]{Guilherme S.~Rodrigues}
  \affil[1]{University of Brasília}
  \affil[2]{University of New South Wales, Sydney}
  \affil[3]{National University of Singapore}
  \maketitle
} \fi

\if1\blind
{
  \bigskip
  \bigskip
  \bigskip
  \begin{center}
    {\LARGE\bf Title}
\end{center}
  \medskip
} \fi

\bigskip
\begin{abstract}
Positional Encoder Graph Neural Networks (PE-GNNs) are among the most effective models for learning from continuous spatial data. However, their predictive distributions are often poorly calibrated, limiting their utility in applications that require reliable uncertainty quantification. We propose the Positional Encoder Graph Quantile Neural Network (PE-GQNN), a novel framework that combines PE-GNNs with Quantile Neural Networks, partially monotonic neural blocks, and post-hoc recalibration techniques. The PE-GQNN enables flexible and robust conditional density estimation with minimal assumptions about the target distribution, and it extends naturally to tasks beyond spatial data. Empirical results on benchmark datasets show that the PE-GQNN outperforms existing methods in both predictive accuracy and uncertainty quantification, without incurring additional computational cost. We also provide theoretical insights and identify important special cases arising from our formulation, including the PE-GNN.
\end{abstract}


\section{Introduction}
\label{sec:intro}

Large spatial datasets are naturally generated in a wide range of applications in economics \citep{anselin2022}, meteorology \citep{bi2023}, urban transportation \citep{lv2014, derrow2021, kashyap2022}, social networks \citep{xu2020}, e-commerce \citep{sreenivasa2019} and other fields. Gaussian Processes (GPs) \citep{rasmussen+w06, cressie+w11} are a fundamental tool for modelling spatial data on continuous domains. They are flexible and interpretable models for unknown functions, both in spatial and more general regression settings. However, with time complexity $O(n^3)$ and storage complexity $O(n^2)$, naive GP methods quickly become intractable for large datasets. This has led to a large range of approximate inference methods, such as those based on sparse approximations to covariance or precision matrices \citep{furrer+gn06, lindgren+rl11}, low rank approximations \citep{cressie+sz22} or nearest neighbour approximations \citep{vecchia98, datta2016, katzfuss+g21}.

Given the difficulty of GP computations, it is of interest to explore scalable methods for large spatial datasets using neural networks (NNs) and to enhance their ability to quantify uncertainty. A prominent method for making spatial predictions using Graph Neural Networks (GNNs) is the Positional Encoder Graph Neural Network (PE-GNN) of \cite{klemmer2023}. Our contribution is to make key modifications to the PE-GNN architecture to enhance its ability to make accurate spatial predictions and to quantify uncertainty. These modifications will be explained further below.

NNs are popular in data modeling and prediction tasks like computer vision and natural language processing (NLP). However, traditional NNs struggle to handle spatial dynamics or graph-based data effectively. GNNs \citep{kipf2017, velickovic2018, hamilton2017} offer a powerful and scalable method for applying NNs to graph-structured data. The idea is to share information through the edges of a graph, allowing nodes to exchange information during learning. GNNs are versatile and can uncover nonlinear relationships among inputs, hidden layers, and each node's neighborhood information. The success of GNNs in spatial applications largely depends on the spatial graph construction, including choice of distance metric and the number of neighboring nodes, and traditional GNNs often struggle to model complex spatial relationships. To address this, \cite{klemmer2023} introduced the PE-GNN, which enhances predictive performance in spatial interpolation and regression. However, the PE-GNN is not designed to provide a full probabilistic description of the target's distribution, and assuming a Gaussian distribution for predictions can lead to poorly calibrated intervals, such as 80\% intervals that fail to contain the true outcome 80\% of the time. Recently, \citet{bao2024} proposed a new framework called Spatial Multi-Attention Conditional Neural Processes (SMACNPs) for spatial small sample prediction tasks. SMACNPs use GPs parameterized by NNs to predict the target variable distribution, which enables precise predictions while quantifying the uncertainty of these predictions. However, these methods remain constrained to producing Gaussian predictive distributions, limiting their capacity to represent asymmetric or multimodal random variates.

Methods based on quantile regression are an alternative approach to probabilistic forecasting making rapid progress in recent years. \citet{Tagasovska2019} introduced the Simultaneous Quantile Regression (SQR) loss function that we use in our formulation. \citet{si2022} proposed a novel architecture for estimating generic quantiles of a conditional distribution. In one dimension, this method produces a quantile function regression $\hat{q}(\tau)|\bm{x}$ that estimates the $\tau$-th quantile of the predictive distribution of the target variable given a feature vector $\bm{x}$. \citet{kuleshov2022} argue that the method of \citet{si2022} is inefficient with high-dimensional predictors. To address this, they modify the original formulation to incorporate a post hoc recalibration procedure whereby an auxiliary model recalibrates the predictions of a trained model. The first model outputs features, usually summary statistics like quantiles, representing a low-dimensional view of the conditional distribution. The auxiliary model, the recalibrator, uses these features as input to produce calibrated predictions using Si {\it et al.}'s quantile function regression framework. The main drawback is that it requires training two separate models, each needing its own training set.

Our work makes two sets of contributions. (1) We propose a new architecture that merges the two-step procedure of \citet{kuleshov2022} into a single model by postponing the concatenation of the $\tau$ value used by \citet{si2022}. Additionally, $\tau$ is transformed beforehand to facilitate learning and provide better control over the form of the predictive distribution. To ensure the model outputs valid, non-crossing quantile functions, we employ the partially monotonic blocks introduced by \citet{nolte2023}. This approach enhances the network’s ability to model uncertainty. The model becomes more robust to high-dimensional predictor spaces, even though few assumptions are made about the form of the target's conditional distribution. As a result, a single model can fully describe the predictive conditional distribution and generate quantile predictions and prediction intervals as byproducts. This method can be applied broadly, not just in spatial regression or GNN contexts. We show how to integrate this strategy into the PE-GNN framework to create an intrinsically calibrated model with no extra computational cost. (2) We introduce structural changes to the PE-GNN. Instead of applying the GNN operator to the concatenation of the nodes' features and the spatial embedding, we apply it only to the features. In the PE-GNN, the GNN operator uses neighbours' features to create new node representations but does not include the target value of neighboring nodes, therefore leaving valuable information underexploited. To address this, we introduce the mean target value of a node's neighbours as a feature after the GNN layers, closer to the output. 

The structure of this work is as follows: Section \ref{sec:background} offers a brief background overview, Section \ref{sec:method} outlines the proposed method for geographic data prediction, Section \ref{sec:experiments} shows experimental results on three real-world datasets, and Section \ref{sec:conclusion} concludes.

\section{Background}
\label{sec:background}

\vspace{.2cm}
\textbf{Graph Neural Networks:}
\hspace{.15cm}
Graph Neural Networks (GNNs) are powerful and scalable tools for representation learning and inference on graph-structured data. They exploit the topological relationships between adjacent nodes to produce context-aware embeddings, which can be effectively used in downstream tasks \citep{wu2022}. GNN layers iteratively refine each node’s embedding by aggregating information from its own features as well as those of its neighboring nodes.

Graph Convolutional Networks (GCNs)~\citep{kipf2017} are a specific type of GNN layer inspired by the convolution operations used in Convolutional Neural Networks (CNNs). For weighted graphs, a GCN layer $k$ updates node embeddings according to the following equation:
\begin{equation}
    \bm{H}^{(k)} = f^{(k)} \left( \bm{D}^{-1/2} \left[ \bm{A} + \bm{I} \right] \bm{D}^{-1/2} \bm{H}^{(k-1)} \bm{W}^{(k)} \right), \quad \text{for } k \in \{1, \ldots, K\}.
\end{equation}
In this formulation, the input to the network is the feature matrix $\bm{H}^{(0)} = \bm{X}$. The function $f^{(k)}$ denotes a non-linear activation function (e.g., ReLU), and $\bm{W}^{(k)}$ is a learnable weight matrix. The adjacency matrix $\bm{A}$ encodes the graph structure, with edge weights typically derived from node distances and zero entries for unconnected pairs. The identity matrix $\bm{I}$ adds self-loops, and $\bm{D}$ is the corresponding \emph{degree} matrix. In our experiments (Section~\ref{sec:experiments}), we also consider other widely used GNN architectures, namely, Graph Attention Networks (GATs)~\citep{velickovic2018} and GraphSAGE~\citep{hamilton2017}.

\vspace{.2cm}
\textbf{Positional Encoder Graph Neural Network:}
\hspace{.15cm}
In a typical spatial regression setting, each datapoint is represented as $p_i = \{y_i, \bm{x}_i, \bm{c}_i\}$, where $y_i$ is a continuous scalar target variable, $\bm{x}_i$ is a vector of input features, and $\bm{c}_i$ denotes the geographical coordinates associated with observation $i$. A given batch of datapoints $B=\{p_1, \ldots, p_{n_B}\}$ can be fully represented by three matrices: the target vector $\bm{y}_B \in \mathbb{R}^{n_B \times 1}$, the feature matrix $\bm{X}_B \in \mathbb{R}^{n_B \times p}$, and the coordinate matrix $\bm{C}_B \in \mathbb{R}^{n_B \times 2}$, respectively.

\citet{klemmer2023} introduced a novel approach for incorporating spatial information into GNNs: the Positional Encoder Graph Neural Network (PE-GNN). In this framework, a positional encoder (PE) processes the spatial coordinate matrix $\bm{C}_B$ to produce a learned spatial embedding matrix $\bm{C}_B^{\text{emb}}$. The embedding is computed as
$\bm{C}_B^{\text{emb}} = PE(\bm{C}_B, \sigma_{\min}, \sigma_{\max}, \Theta_{\text{PE}}) = NN(ST(\bm{C}_B, \sigma_{\min}, \sigma_{\max}), \Theta_{\text{PE}})$, where $ST$ denotes a deterministic set of sinusoidal transformations with hyperparameters $\sigma_{\min}$ and $\sigma_{\max}$, and $NN$ is a fully connected neural network with trainable parameters $\Theta_{\text{PE}}$. A complete description of the PE mechanism is provided in Appendix~\ref{sub:PE}.

The matrix $\bm{C}_B^{\text{emb}}$ is concatenated column-wise with the node features before the application of GNN layers. Thus, the input to the first GNN layer is given by
$\bm{H}^{(0)}_B = \text{concat}(\bm{X}_B, \bm{C}^{\text{emb}}_B).$
At each training step, a new random batch of nodes $B$ is sampled, and the full pipeline --- graph construction, spatial embedding generation, feature concatenation, and GNN propagation --- is executed using only the nodes in that batch. For each node $p_i \in \{p_1, \ldots, p_{n_B}\}$, the PE-GNN predicts a target value $\hat{y}_i$ and, as an auxiliary task~\citep{klemmer2021}, the corresponding \emph{Local Moran’s I} statistic~\citep{anselin1995}, denoted by $\hat{I}(y_i)$. The total loss used by \citet{klemmer2023} combines both objectives:
\[ \mathcal{L}_B = \text{MSE}(\hat{\bm{y}}_B, \bm{y}_B) + \lambda \, \text{MSE}(\hat{I}(\bm{y}_B), I(\bm{y}_B)), \] where $\lambda$ controls the contribution of the auxiliary task.

\vspace{.2cm}
\textbf{Quantile regression:}
\hspace{.15cm}
\citet{koenker1978} proposed a linear quantile regression model to estimate conditional distribution quantiles. It uses the pinball loss $\rho_{\tau} (r_i) = \max{(\tau r_i, (\tau - 1) r_i)}$, where $r_i = y_i - \hat{q}_i(\tau)$, $\hat{q}_i(\tau) = \bm{X}_i \hat{\bm{\beta}}$, and $\tau$ is the desired cumulative probability associated with the predicted quantile $\hat{q}_i(\tau)$. The pinball loss for the $i$-th observation is $\rho_{\tau} (r_i)$. The loss over a dataset is the average $\rho_{\tau} (r_i)$ value over all datapoints. A natural extension of quantile linear regression is quantile neural networks (QNNs). This approach is illustrated in Figure \ref{fig:nonlinear}, which seeks to estimate the conditional quantiles for a pre-defined grid $(\tau^1, \ldots, \tau^d)$. Each quantile is estimated by an independent model (Figure \ref{fig:nonlinear}). This can lead to quantile predictions with quantile crossing (e.g., a median prediction lower than the first quartile prediction). \citet{rodrigues2020} proposed an approach that outputs multiple predictions: one for the expectation and one for each quantile of interest. The loss function is:
\begin{equation}
    \label{eq:pinball-multi}
    \mathcal{L} = \frac{1}{d + 1} \left[ \text{MSE} \left( \hat{\bm{y}}, \bm{y} \right) + \sum_{i=1}^{n} \sum_{j=1}^{d} \frac{\rho_{\tau^j} \left(y_i - \hat{q}_i(\tau^j) \right)}{n} \right].
\end{equation}

\citet{Tagasovska2019} proposed a method to generate a model that is independent of quantile selection. During training, for each datapoint in the batch, a Monte Carlo sample $\tau \sim U(0, 1)$ is drawn and concatenated with the corresponding datapoint feature vector. The SQR loss function is similar to Eqn.~\ref{eq:pinball-multi}, but they predict random quantiles $\mathcal{L} = \frac{1}{n} \sum_{i=1}^{n} \rho_{{\tau}_i} ( y_i - \hat{q}_i({\tau}_i) )$, with $\{\tau_i\}_1^n \sim U(0, 1) $. As the network learns, it becomes able to provide a direct estimate to \textit{any} quantile of interest. Hence, this procedure outputs an inherently calibrated model suitable for conditional density estimation. \citet{si2022} construct NNs with a similar loss function (Figure \ref{fig:si}). 

\citet{kuleshov2022} adapted the architecture from \citet{si2022} into a two-step process for larger predictor spaces (Figure \ref{fig:kuleshov}). First, a model is trained to take the original features as inputs and generate low-dimensional representations of the predicted distribution. Next, a recalibrator is trained using  these \emph{new} features by minimizing the estimated expected pinball loss over $\tau$. During inference, the recalibrator takes the new features and an arbitrary $\tau$ as inputs to produce the quantile prediction. This method is highly dependent on the choice of recalibrator features.

\section{Method}
\label{sec:method}

In this work, we introduce the \textbf{Positional Encoder Graph Quantile Neural Network (PE-GQNN)}, a novel framework for predictive modeling on spatial data. Algorithm \ref{alg:alg1} shows the step-by-step procedure to train a \textbf{PE-GQNN} model. 

\begin{figure}[t]
    \centering
    \includegraphics[width=\textwidth]{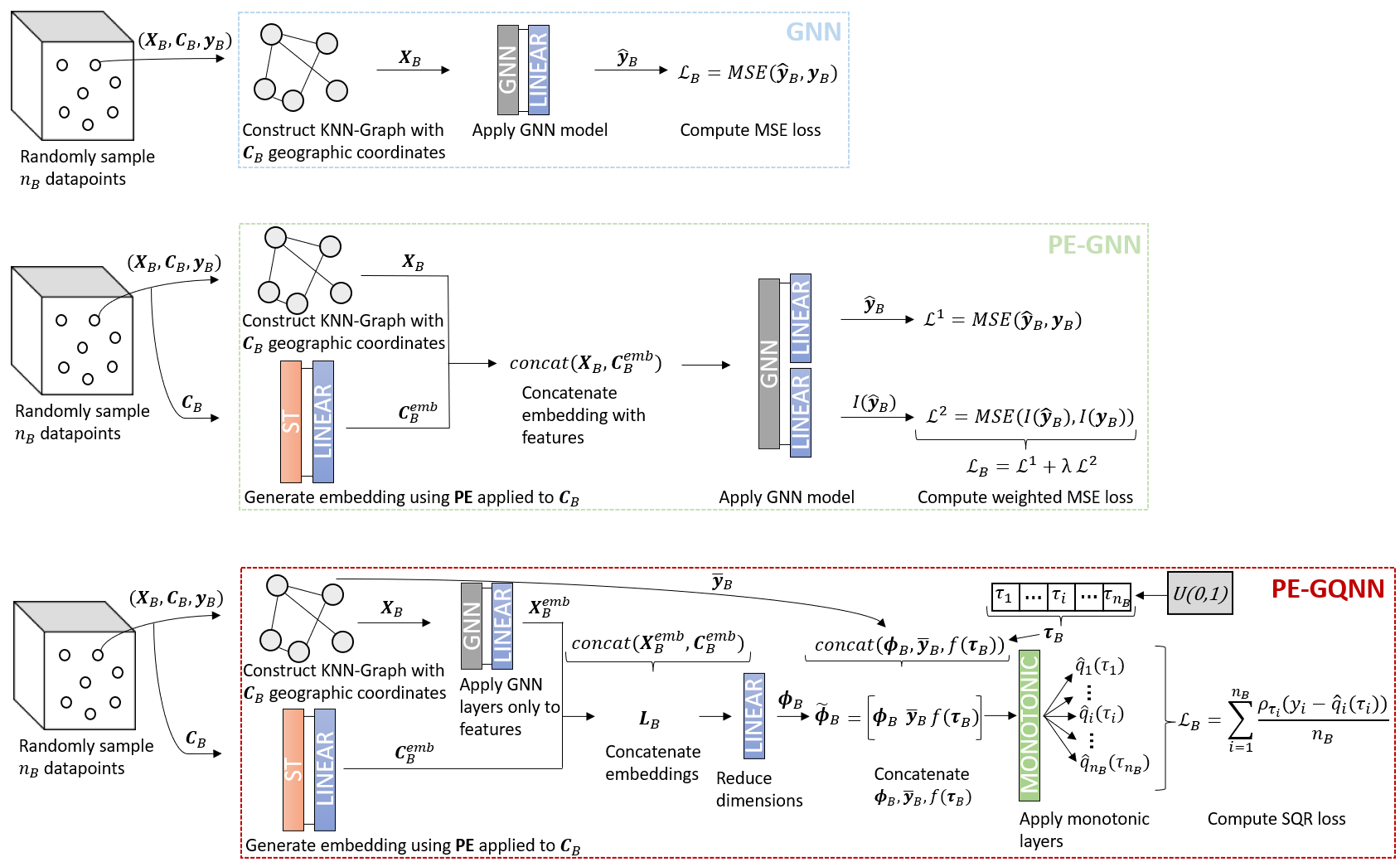}
    \caption{\textbf{PE-GQNN} compared to PE-GNN and GNN}
    \label{fig:draw}
\end{figure}

Figure \ref{fig:draw} illustrates its complete pipeline. Here, each rectangle labeled "GNN", "LINEAR" and "MONOTONIC" represents a set of one or more neural network layers, with the type of each layer defined by the title inside the rectangle. At each layer, a nonlinear transformation (e.g.~ReLU) may be applied.

\begin{algorithm}[H]
\caption{PE-GQNN training}
\begin{algorithmic}[1] 
\Require 
\Statex Training data target, features, and coordinates matrices: ${\bm{y}}_{(n \times 1)}$, ${\bm{X}}_{(n \times p)}$, and ${\bm{C}}_{(n \times 2)}$.
\Statex A positive integer $k$ defining the number of neighbors considered in the spatial graph.
\Statex Positive integers $tsteps$ and $n_B$, the number of training steps and the batch size.
\Statex Positive integers $u$, $g$, and $s$, the embedding dimensions considered in, respectively, the PE, the GNN layers, and the layer where we introduce $\bm{\tau}$ and $\bar{\bm{y}}$.
\Statex An activation function $f(\phantom{})$ for $\bm{\tau}$.

\Ensure
\Statex A set of learned weights for the model initialized at Step \ref{alg1-random}.
\Statex

\State Initialize model with random weights and hyperparameters. \label{alg1-random}
\State Set optimizer with hyperparameters. \vspace{.2cm}

\For{$b \gets 1$ \textbf{to} $tsteps$} \Comment{Batched training}
    \State Sample minibatch $B$ of $n_B$ datapoints: ${\bm{X}_B}_{(n_B \times p)}$, ${\bm{C}_B}_{(n_B \times 2)}$, ${\bm{y}_B}_{(n_B \times 1)}$.
    \State Input ${\bm{C}_B}_{(n_B \times 2)}$ into PE, which outputs the batch's spatial embedding matrix ${\bm{C}_B^{emb}}_{(n_B \times u)}$. \label{alg:alg1-pe}
    \State Compute the great-circle distance between each pair of datapoints from ${\bm{C}_B}$. \label{alg:alg1-dist}
    \State Construct a graph using $k$-nearest neighbors from the distances computed in Step \ref{alg:alg1-dist}. \label{alg:alg1-graph}
    \State Set ${\bm{A}_B}$ as the adjacency matrix of the graph constructed in Step \ref{alg:alg1-graph}.
    \For{$i \gets 1$ \textbf{to} $n_B$}
        \State Using ${\bm{A}_B}$, compute $\bar{y}_{i} = \frac{1}{k} \sum_{j=1}^{k} y_{j}$, where $j = {1, \ldots, k}$ are the neighbors of $i$. \label{alg:alg1-knn}
    \EndFor
    \State Set $\bar{\bm{y}}_{B} = \left[ \bar{y}_{1}, \ldots, \bar{y}_{n_B} \right]^{\top}$.
    \State Apply GNN layers to the features ${\bm{X}_B}_{(n_B \times p)}$, followed by fully-connected layers to reduce dimensionality. This step outputs a feature embedding matrix ${\bm{X}_B^{emb}}_{(n_B \times g)}$. \label{alg:alg1-femb}
    \State Column concatenate ${\bm{X}_B^{emb}}_{(n_B \times g)}$ with ${\bm{C}_B^{emb}}_{(n_B \times u)}$, which results in ${\bm{L}_B}_{(n_B \times (g + u))}$. \label{alg:alg1-concat}
    \State Apply fully-connected layers to reduce ${\bm{L}_B}_{(n_B \times (g + u))}$ to ${\bm{\phi}_B}_{(n_B \times s)}$. \label{alg:alg1-reduce}
    \State Create a vector with values sampled from $U(0, 1)$: ${\bm{\tau}_B}_{(n_B \times 1)} = \left[ \tau_1, \ldots, \tau_{n_B} \right]^{\top}$. \label{alg:alg1-uni}
    \State Column concatenate $\bm{\phi}_B$ with $f\left(\bm{\tau}_B\right)$ and $\bar{\bm{y}}_{B}$ to create $\bm{\widetilde{\phi}}_B{}_{(n_B \times (s + 2))}$. \label{alg:alg1-intro}
    \State Predict the target quantile vector $\left[ \hat{q}_1(\tau_1), \ldots, \hat{q}_{n_B}(\tau_{n_B}) \right]^{\top}$ using $\widetilde{\bm{\phi}}_B$. \label{alg:alg1-predict}
    \State Compute loss $\mathcal{L}_B = \frac{1}{n_B} \sum_{i=1}^{n_B} \rho_{\tau_i} \left( y_i - \hat{q}_i(\tau_i) \right)$.
    \State Update the parameters of the model using stochastic gradient descent.
\EndFor \vspace{.1cm}
\end{algorithmic}
\label{alg:alg1}
\end{algorithm}

After initializing the model and hyperparameters, the first step of \textbf{PE-GQNN} is to randomly sample a batch $B$ of datapoints. The next step projects the matrix of geographical coordinates $\bm{C}_B$ into the positional embeddings, ${\bm{C}_B^{emb}}_{(n_B \times u)}$ (Algorithm \ref{alg:alg1}, Step \ref{alg:alg1-pe}). $\bm{C}_B$ is also used to compute the distance between each pair of datapoints (Step \ref{alg:alg1-dist}). From these distances and a predefined number of nearest neighbors, a graph can be constructed, with each datapoint as a node and edge weights computed from the distances, leading to the batch adjacency matrix $\bm{A}_B$.

At Step \ref{alg:alg1-femb}, the first distinction between \textbf{PE-GQNN} and PE-GNN arises: instead of using the concatenation of the feature matrix and the spatial embedding as the input for the GNN operator, we apply the GNN operator only to the feature matrix $\bm{X}_B$. One or more fully-connected layers are then used to reduce the feature embedding dimensionality. This process receives the constructed graph and the batch feature matrix ${\bm{X}_B}_{(n_B \times p)}$ as input and yields an embedding matrix of features as output: ${\bm{X}_B^{emb}}_{(n_B \times g)}$. Step \ref{alg:alg1-concat} performs a column concatenation between the feature embedding ${\bm{X}_B^{emb}}_{(n_B \times g)}$ and the output obtained from the PE: ${\bm{C}_B^{emb}}_{(n_B \times u)}$. This concatenation results in the matrix ${\bm{L}_B}_{(n_B \times (g + u))}$. 

Subsequently, we use one or more fully-connected layers (Step \ref{alg:alg1-reduce}) to reduce the dimensionality of $\bm{L}_B$, making it suitable for two innovations in \textbf{PE-GQNN}. This set of fully-connected layers outputs the matrix ${\bm{\phi}_B}_{(n_B \times s)}$, which is then combined with $\bar{\bm{y}}_{B}$ and $\bm{\tau}_B$. $\bar{\bm{y}}_{B}$ represents a vector with one scalar for each datapoint in the batch, containing the mean target variable among the training neighbours for each node. It is computed using the graph constructed in previous steps (Step \ref{alg:alg1-knn}), and has dimensions $n_B \times 1$. It is comparable to a vector of predictions generated by a KNN regression model, where neighbours are determined using the distance calculated from geographical coordinates. Here, we used the simple average due to its relationship with KNN prediction; however, one could use a weighted average via the adjacency matrix $\bm{A}_B$. We introduce this input at a later stage to avoid data leakage. If the GNN operator received $\bar{\bm{y}}_{B}$ as input, after completing the message passing process in each GNN layer, the true node target value would inadvertently be transmitted to its neighbours, creating potential data leakage \citep{appleby2020}.

In the same layer where $\bar{\bm{y}}_{B}$ is introduced, we apply a similar approach to \citet{si2022} to make \textbf{PE-GQNN} an inherently calibrated model suitable for probabilistic and quantile predictions. For each batch $B$, we create a $n_B \times 1$
vector ${\bm{\tau}_B}_{(n_B \times 1)} = [ \tau_1, \ldots, \tau_{n_B} ]^{\top}$ of random $U(0,1)$ draws (Step \ref{alg:alg1-uni}). Then, we column concatenate $\bm{\phi}_B$ with $f\left(\bm{\tau}_B\right)$ and $\bar{\bm{y}}_{B}$ to create $\bm{\widetilde{\phi}}_B{}_{(n_B \times (s + 2))}$ (Step \ref{alg:alg1-intro}), where $f(\phantom{})$ is an activation function. Next, forward propagation is computed (Step \ref{alg:alg1-predict}) in one or more fully-connected layers, outputting predicted quantiles for each datapoint in the batch. These predictions are then used to compute the SQR loss function introduced by \citet{Tagasovska2019}.

Incorporating $\tau$ values into the model architecture improves its ability to model uncertainty and serves as a regularization mechanism \citep{rodrigues2020}. The use of SQR loss acts as a natural regularizer, producing a detailed description of the predictive density beyond just mean and variance estimation. For predictions, the quantile of interest, $\tau$, must be given, along with the basic data components (e.g. $\tau=0.25$ gives the first quartile). If interest is in predicting multiple quantiles for the same observation, the input can be propagated up to the layer where $\bm{\tau}$ is introduced. For each quantile of interest, propagation can be limited to the final layers.

\vspace{.1cm}
\textbf{Homoscedastic Gaussian model as a particular case:}
\hspace{.15cm}
If in Steps \ref{alg:alg1-intro} and \ref{alg:alg1-predict} one sets $f(\tau) = \Phi^{-1}(\tau)$, where $\Phi()$ denotes the PDF of a Standard Gaussian distribution, and use a single linear predictive layer, the model's quantile function,
\begin{equation}
    \label{eq:quantile-crossing}
    \hat{q}_i(\tau) = (b + w_{\bar{y}_i} \bar{y}_i + \sum_{j=1}^{s} w_j \bm{\phi}_j ) + w_{\tau} \Phi^{-1}(\tau) = \mu_i + \sigma \Phi^{-1}(\tau), \text{ } \forall i \in {1, \ldots, n_B},
\end{equation}
matches the quantile function of a Gaussian random variable, therefore reducing \textbf{PE-GQNN} to a homoscedastic Gaussian regression model with mean $\mu_i$ and a common, learnable standard deviation $\sigma = w_{\tau}$. Here, $b$, $w_{\tau}$, $w_{\bar{y}_i}$, and $\{w_j\}$ are the prediction layer parameters, and $\{\bm{\phi}_j\}$ are the activation values from the previous layer. This result may appear counterintuitive, as the SQR loss can produce a model structurally identical to one trained with a fundamentally different objective, namely, the MSE loss. Similarly, if $f(\tau) = \text{logit}(\tau)$ and a single linear predictive layer is employed, the model yields logistic predictive distributions, a continuous distribution that should not be confused with logistic regression. Therefore,  a well-chosen activation function $f(\tau)$ not only facilitates learning but also explicitly shapes the structure of the predictive distributions. The use of monotonic layers enhances the flexibility of our approach, significantly reducing the reliance on the choice of $f(\tau)$.

\vspace{.1cm}
\textbf{Target domain:}
\hspace{.15cm}
The final layer should preferably use an activation function coherent with the domain of the target variable, ensuring model outputs are valid for target distribution support. E.g., an exponential function could be appropriate if the target variable is continuous, unbounded and positive.

\vspace{.1cm}
\textbf{Quantile crossing and Monotonic Blocks:}
\hspace{.15cm}
Several related approaches \citep{Tagasovska2019, rodrigues2020, si2022, kuleshov2022} are subject to quantile crossing, a phenomenon that occurs when the requirement that higher quantiles be greater than or equal to lower quantiles is violated. To ensure the network outputs valid probability distributions (i.e. $\hat{q}(\tau, \widetilde{\bm{\phi}}, \bar{y}) \leq \hat{q}(\tau', \widetilde{\bm{\phi}}, \bar{y}), \forall \, \tau < \tau'$), we propose using the Lipschitz Monotonic Networks (LMN) introduced by \citet{nolte2023} to approximate $q(\tau, \widetilde{\bm{\phi}}, \bar{y})$. Those are highly expressive blocks of layers that can approximate all monotonic Lipschitz bounded functions. That is, any function for which there exists a constant $\lambda$ such that $|\hat{q}(\tau, \widetilde{\bm{\phi}}, \bar{y}) - \hat{q}(\tau', \widetilde{\bm{\phi}}, \bar{y})| \leq \lambda |\tau - \tau'|, \; \forall  \; 0 < \tau , \tau' < 1$.

\section{Experiments}
\label{sec:experiments}

\subsection{Experimental setup}
\label{subsec:setup}

The \textbf{PE-GQNN} was implemented using PyTorch \citep{pytorch2019} and PyTorch Geometric \citep{pyg2019}. 
We conducted comprehensive simulations to explore the prediction performance and other properties of the proposed model. Computation was performed on an Intel i7-7500U processor with 16 GB of RAM, running Windows 10.

\vspace{.2cm}
\textbf{Candidate models:}
\hspace{.15cm}
The experiment was designed to compare five primary methods approaches for addressing spatial regression problems across three diverse real-world datasets (California Housing, Air Temperature, 3D Road). Table \ref{tab:summary} lists each candidate model and their applicable datasets. All models were trained using the Adam optimizer \citep{kingma2015}, with early stopping employed to prevent overfitting. For GNN-based approaches, we used $k = 5$ nearest neighbors to construct the graphs. The learning rate was set to 0.001 across all models. A batch size of 1,024 was used for the Air Temperature dataset, while a batch size of 2,048 was used for the remaining two datasets. The architectural details are given in Appendix Section \ref{sub:details}.

\begin{table}[!htbp]
\centering{
    \caption{Summary of candidate models.}
    \label{tab:summary}
}
\resizebox{\textwidth}{!}{
\begin{tabular}{l|l|l|l|l|l|l}
\textbf{Approach} & \textbf{Model} & \textbf{Type} & \textbf{PE} & \textbf{Innovations} & \textbf{Loss} & \textbf{Datasets} \\ \hline \hline	
I & GNN & GNN & No & No & MSE & All \\
II & PE-GNN $\lambda = \text{best}$ & GNN & Yes & No & $\text{MSE}_y + \lambda \text{MSE}_{I(y)}$ & All \\
III & PE-GNN (with SQR) & GNN & Yes & No & SQR & California \\
IV & PE-GQNN & GNN & Yes & Yes & SQR & All \\
V & SMACNP & GP & No & No & Log Likelihood & All but 3D Road \\
\end{tabular}
}
\end{table}

Approach I involves the traditional application of GNNs to geographic data. Three types of GNN layers were considered: GCNs \citep{kipf2017}, GATs \citep{velickovic2018}, and GSAGE \citep{hamilton2017}. For each of these, the architecture remains consistent to facilitate performance comparisons: two GCN/GAT/GSAGE layers with ReLU activation and dropout, followed by a linear prediction layer.

Approach II involves the application of PE-GNN \citep{klemmer2023} with optimal weights for each dataset and layer type combination, as demonstrated by the experimental findings of \citet{klemmer2023}. The GNN architecture used is the same as for approach I. It was implemented using the code available at: \url{https://github.com/konstantinklemmer/pe-gnn}.

Approach III represents a naive combination of the PE-GNN with the quantile regression framework described in Section \ref{sec:method}. Specifically, we trained the PE-GNN with the SQR loss function, concateneting $f(\tau)=\tau$ immediately after the GNN layers. Approach IV, which is the primary focus of this research, is the \textbf{PE-GQNN}. Compared to Approach III, it introduces the following innovations: $\tau$ is incorporated into the network through $\text{probit}(\tau)$, but only after reducing $L_B$ into $\phi_B$. The GNN layers no longer process the positional encoders, and the final fully connected layers are replaced with monotonic blocks. Additionally, $\bar{y}$ is used as a feature in the final part of the network. The architectures of the PE and GNN layers remain identical to those in the previous approaches.

Finally, a benchmark approach that does not use GNNs but was recently proposed for modelling spatial data will be considered as approach V: SMACNPs. This approach, proposed by \citet{bao2024}, has demonstrated superior predictive performance, surpassing GPs models in the three real-world datasets considered. This model was implemented following the specifications of \citet{bao2024}, using the code available at: \url{https://github.com/bll744958765/SMACNP}.

Approaches I and II do not inherently provide predicted conditional distributions. However, as they optimize the MSE metric, they implicitly learn a Maximum Likelihood Estimate (MLE) of a Gaussian model. Thus, the predictive distribution considered for these approaches was a Gaussian distribution centered on the point prediction with variance equal to the MSE of the validation set. For computational simplicity in the experiments, instead of calculating $\bar{\bm{y}}_{B}$ for each batch, we pre-calculated $\bar{\bm{y}}$ using the entire training set.

\vspace{.2cm}
\textbf{Performance metrics:}
\hspace{.15cm}
We evaluate predictive accuracy using Mean Squared Error (MSE) and Mean Absolute Error (MAE). To assess calibration of the predictive distributions, we report the SQR loss and calibration metric introduced by \citet{kuleshov2018}: 

\begin{equation*}
\text{calibration} = \sum_{j=1}^{m} \left ( \tau^j - \frac{1}{n} \sum_{i=1}^{n} \mathds{1}\left [ y_i \leq \hat{q}_i(\tau^j) \right ] \right )^2. 
\end{equation*}

For quantile predictions of a calibrated model for a given $\tau$, the proportion of observed values less than or equal to the predicted quantile should approximate $\tau$. Evaluating the calibration metric helps determine whether the predicted quantiles are accurate.

\subsection{California Housing}
\label{subsec:california}

This dataset comprises pricing information for $>$20,000 residential properties in California, recorded during the 1990 U.S. census \citep{pace1997}. The main objective is a regression task: predict housing prices, $y$, through the incorporation of six predictive features, $\bm{x}$, and geographical coordinates, $\bm{c}$. The predictive features are neighborhood income, house age, number of rooms, number of bedrooms, occupancy and population. All models were trained and evaluated using 80\% of the data for training, 10\% for validation, and 10\% for testing. In the case of SMACNP, to adhere to the specifications of \citet{bao2024}, a training subsample was extracted to represent 10\% of the entire dataset.  The training time, number of training epochs, number of parameters, and final performance metrics on the test dataset are summarized in Table \ref{tab:results-cali}.

\begin{figure}[t]
     \centering
     \begin{subfigure}[b]{0.32\textwidth}
         \centering
         \includegraphics[width=\textwidth]{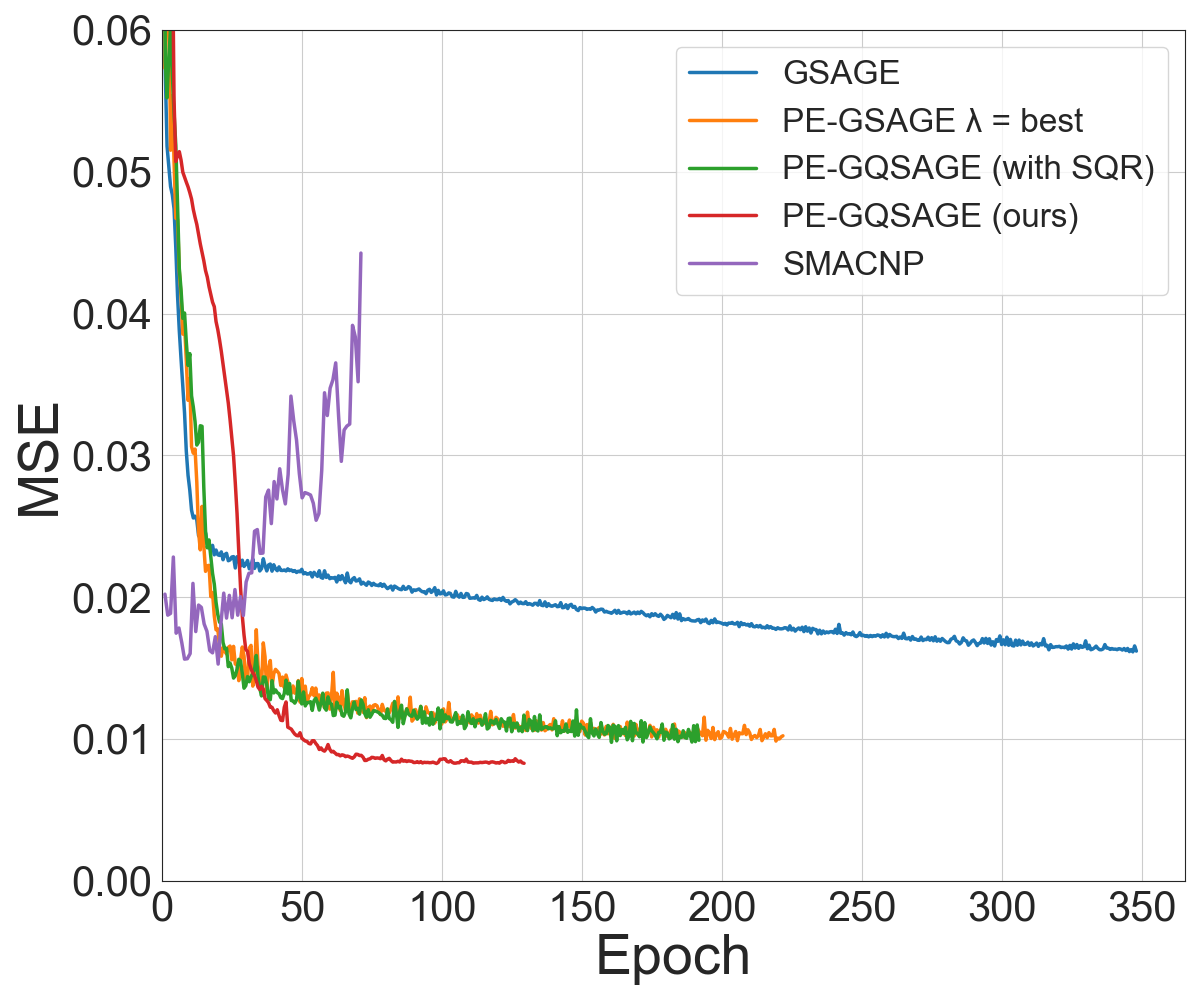}
         \caption{Validation MSE.}
         \label{fig:val-mse-cali-gsage}
     \end{subfigure}
     \begin{subfigure}[b]{0.32\textwidth}
      \centering
         \includegraphics[width=\textwidth]{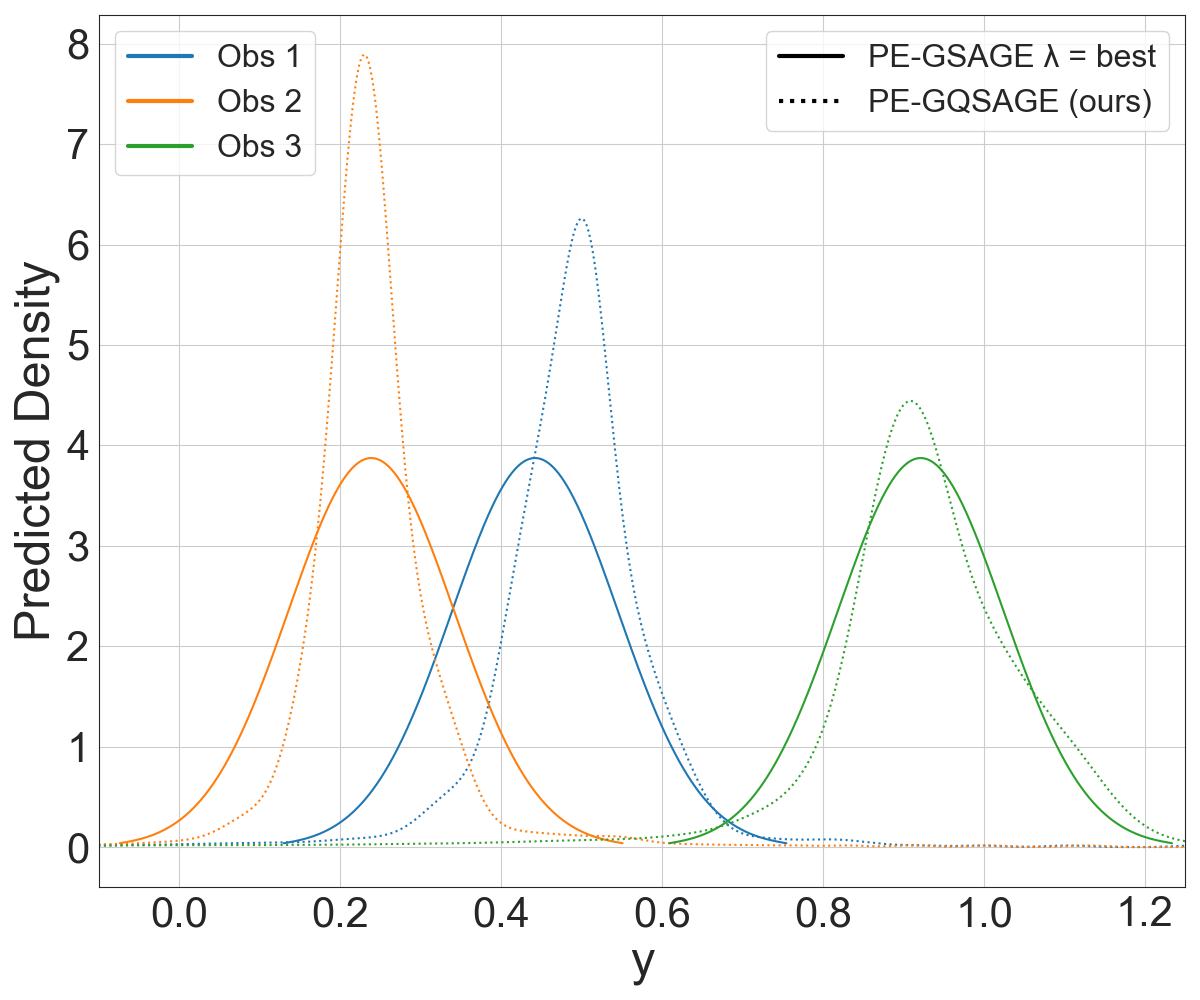}
         \caption{Predictive densities.}
         \label{fig:cali-density}
     \end{subfigure}
     \begin{subfigure}[b]{0.32\textwidth}
         \centering
         \includegraphics[width=\textwidth]{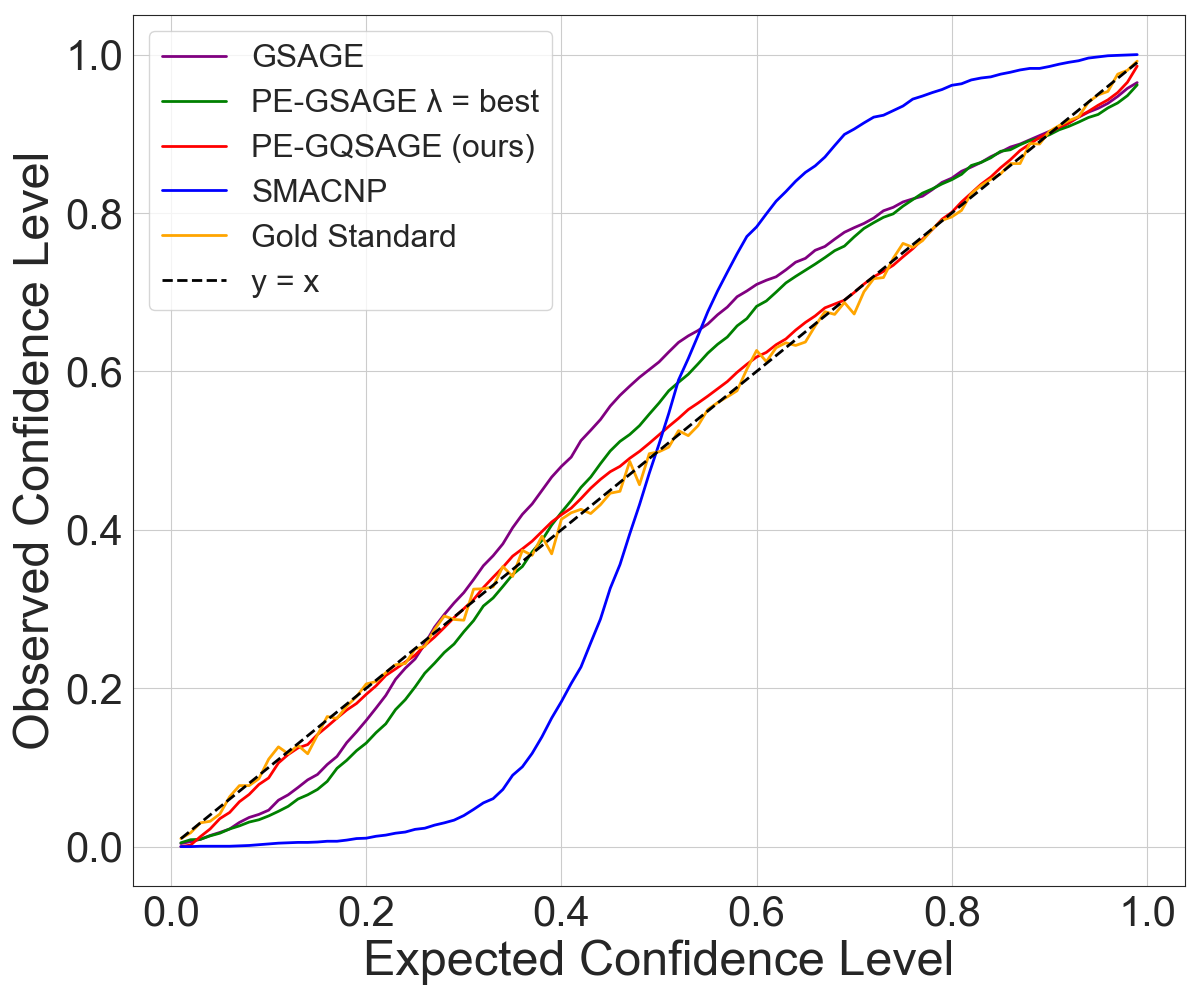}
         \caption{Calibration Plot.}
         \label{fig:cali-ecp}
     \end{subfigure}
     \caption{Diagnostics for the California Housing dataset. (a) Validation MSE curves. (b) PE-GQSAGE predicted densities of 3 observations of the test set. (c) Calibration Plot.}
     \label{fig:cali-dist}
\end{figure}

\begin{table}[!htbp]
\caption{Performance metrics on the California Housing test set. \emph{Cover.} refers to the observed coverage for 95\% confidence intervals. }
\label{tab:results-cali}
\begin{center}
\resizebox{.95\textwidth}{!}{
\begin{tabular}{l|l|l|l|l|l|l|l|l}
\textbf{Model} & \textbf{Time} & \textbf{Epochs} & \textbf{\# Param.} & \textbf{MSE} & \textbf{MAE} & \textbf{SQR} & \textbf{Cover.} & \textbf{Calibr.} \\ \hline \hline
GCN & 00h41m32s & 441 & \textbf{1,313} & 0.022 & 0.110 & 0.040 & 0.930 & 0.301 \\
PE-GCN $\lambda = \text{best}$ & 00h17m13s & 170 & 24,129 & 0.018 & 0.093 & 0.036 & 0.920 & 0.262 \\
PE-GCN (with SQR) & \textbf{00h12m08s} & 113 & 25,217 & 0.018 & 0.092 & 0.034 & 0.870 & 0.132 \\
PE-GQCN (ours) & 00h14m44s & 129 & 26,201 & 0.011 & 0.069 & 0.026 & 0.927 & 0.118 \\ \hline
GAT & 00h33m29s & 398 & 1,441 & 0.023 & 0.110 & 0.041 & 0.931 & 0.500 \\
PE-GAT $\lambda = \text{best}$ & 00h15m43s & 120 & 24,290 & 0.018 & 0.093 & 0.035 & 0.919 & 0.296 \\
PE-GAT (with SQR) & 00h13m01s & 120 & 25,345 & 0.018 & 0.096 & 0.035 & 0.843 & 0.153 \\
PE-GQAT (ours) & 00h12m49s & 113 & 26,329 & 0.012 & 0.069 & 0.026 & 0.952 & 0.057 \\ \hline
GSAGE & 00h31m13s & 348 & 2,529 & 0.017 & 0.094 & 0.035 & 0.945 & 0.446 \\
PE-GSAGE $\lambda = \text{best}$ & 00h29m02s & 222 & 27,426 & 0.011 & 0.073 & 0.027 & 0.935 & 0.270 \\
PE-GSAGE (with SQR) & 00h20m08s & 192 & 28,481 & 0.011 & 0.071 & 0.026 & 0.834 & 0.126 \\
PE-GQSAGE (ours) & 00h13m39s & 129 & 27,417 & \textbf{0.009} & \textbf{0.060} & \textbf{0.023} & \textbf{0.952} & \textbf{0.016} \\ \hline
SMACNP & 03h15m51s & \textbf{70} & 748,482 & 0.016 & 0.088 & 0.047 & 0.998 & 2.700 \\
\end{tabular}
}
\end{center}
\end{table}

As shown in Table \ref{tab:results-cali}, \textbf{PE-GQNN} delivers strong performance across all evaluation metrics, significantly outperforming traditional GNNs, PE-GNN, and SMACNP. Among models with GSAGE layers --- the overall top performers --- PE-GQSAGE achieved the lowest values for MSE, MAE, SQR, and calibration error. Specifically, compared to PE-GSAGE, it reduced MSE by 23\%, MAE by 18\%, SQR by 14\%, and calibration error by 94\%, all while maintaining the shortest training time. In contrast, a naive combination of PE-GNN with SQR offered no meaningful improvement over PE-GNN alone.

Figure \ref{fig:cali-dist} presents plots that elucidate the behavior of the PE-GQSAGE predictions. The validation MSE curves throughout training are shown in Figure \ref{fig:val-mse-cali-gsage}. Panel \ref{fig:cali-density} illustrates the predicted density of a subsample of three observations from the test set. Parametric models typically assume a fixed output structure --- such as a Gaussian distribution --- which can restrict their expressiveness. In contrast, \textbf{PE-GQNN} imposes minimal assumptions on the form of the predictive distribution, offering greater flexibility. As illustrated in Figure \ref{fig:cali-density}, \textbf{PE-GQNN} is capable of producing predictive distributions with varying shapes and scales, whereas PE-GNN can only differentiate samples based on their location (mean value). It is worth noting that, in this example, assuming Gaussian distributions is reasonable. However, in scenarios involving asymmetric or multimodal distributions, the flexibility of our model-free approach becomes even more advantageous.

Lastly, Figure \ref{fig:cali-ecp} displays the observed confidence level for the test dataset quantile predictions using each of the $m=99$ $\tau$ values in $[0.01, 0.02, \ldots, 0.99]^{\top}$. This type of plot was proposed by \citet{kuleshov2018}. The closer a model gets to the dashed diagonal line, the closer the expected and observed confidence levels are. The Gold Standard represents one Monte Carlo draw from a perfectly specified model, where for each quantile level, the observed confidence level is the observed success rate in $n$ Bernoulli trials with a success probability of $\tau$, where $n$ is the number of test set instances. It is evident that \textbf{PE-GQSAGE} has by far the best calibration performance. This is particularly notable when compared to SMACNP, which exhibits substantial calibration deficiencies due to its tendency to overestimate the variance component.

\subsection{All datasets}
\label{sub:all-datasets}

Experiments were conducted on two other geographic datasets used by \citet{klemmer2023} and \citet{bao2024}. The Air Temperature dataset (\citet{hooker2018}) contains geographical coordinates for $\sim$3,000 meteorological stations worldwide, with the goal of predicting mean temperatures ($y$) using mean precipitation levels ($x$). Models were trained with 80\% of the data, with 10\% for validation and testing each, while SMACNP used a 30\% subsample for training, following the specifications of \citet{bao2024}. The 3D Road dataset (\citet{kaul2013}), includes $>$ 430,000 points with latitude, longitude, and altitude for the Jutland, Denmark road network. The task is to interpolate altitude ($y$) using latitude and longitude ($\bm{c}$). The data were split into 90\% for training, 1\% for validation, and 9\% for testing. SMACNP metrics are not reported due to high computational costs for this dataset.

\begin{table}[!htbp]
\centering{
    \caption{Performance metrics from three different real-world datasets. Models with GSAGE layers.}
    \label{tab:results-others}
}
\resizebox{\textwidth}{!}{
\begin{tabular}{l|l|l|l|l|l|l|l|l|l|l|l|l}
\multirow{2}{*}{\textbf{Model}} & \multicolumn{4}{c|}{\textbf{California Housing}} & \multicolumn{4}{c|}{\textbf{Air Temperature}} & \multicolumn{4}{c}{\textbf{3D Road}} \\
 & \textbf{MSE} & \textbf{MAE} & \textbf{SQR} & \textbf{Calibr.} & \textbf{MSE} & \textbf{MAE} & \textbf{SQR} & \textbf{Calibr.} & \textbf{MSE} & \textbf{MAE} & \textbf{SQR} & \textbf{Calibr.} \\ \hline \hline	
GNN & 0.0170 & 0.0945 & 0.0347 & 0.4456 & 0.0223 & 0.1152 & 0.0427 & \textbf{0.2494} & 0.0170 & 0.1031 & 0.0360 & 0.4384 \\
PE-GNN & 0.0115 & 0.0732 & 0.0270 & 0.2696 & 0.0048 & 0.0526 & 0.0172 & 1.2173 & 0.0031 & 0.0420 & 0.0142 & \textbf{0.1960} \\
PE-GQNN (ours) & \textbf{0.0088} & \textbf{0.0598} & \textbf{0.0231} & \textbf{0.0161} & 0.0033 & 0.0359 & \textbf{0.0140} & 0.3530 & \textbf{0.0001} & \textbf{0.0059} & \textbf{0.0024} & 1.0345 \\ \hline
SMACNP & 0.0160 & 0.0881 & 0.0466 & 2.7000 & \textbf{0.0018} & \textbf{0.0290} & 0.0391 & 5.8611 & \phantom{---} - & \phantom{---} - & \phantom{---} - & \phantom{---} - \\
\end{tabular}
}
\end{table}

Table \ref{tab:results-others} showcases the experimental results obtained from all three datasets: California Housing, Air Temperature, and 3D road. The \textbf{PE-GQNN} models incorporate all innovations discussed in Section \ref{sec:method}. Additionally, we include the SMACNP results as a benchmark model based on GPs \citep{bao2024}. \textbf{PE-GQNN} outperforms both traditional GNN and PE-GNN. Across all datasets, the innovations introduced by \textbf{PE-GQNN} result in substantial reductions in MSE, MAE, and SQR, although, for the 3D Road dataset, PE-GNN achieves the lowest calibration error. In the California Housing dataset, \textbf{PE-GQNN} consistently outperforms SMACNP in predictive accuracy and provides enhanced uncertainty quantification across all types of GNN layers. Conversely, for the Air Temperature dataset, SMACNP achieves the lowest MSE and MAE but suffers from significantly uncalibrated predictions, reflected by a much higher SQR and calibration error compared to \textbf{PE-GQNN}.

\section{Discussion}
\label{sec:conclusion}

In this work, we have proposed the Positional Encoder Graph Quantile Neural Network (\textbf{PE-GQNN}) as an innovative framework to enhance predictive modeling for geographic data. Through a series of experiments on real-world datasets, we have demonstrated the significant advantages of the \textbf{PE-GQNN} over competitive methods. The empirical results underscored the capability of the \textbf{PE-GQNN} to achieve lower MSE, MAE, and SQR across all datasets and GNN backbones compared to traditional GNN and PE-GNN. Notably, the \textbf{PE-GQNN} demonstrated substantial improvements in predictive accuracy and uncertainty quantification. The \textbf{PE-GQNN} framework's ability to provide a full description of the predictive conditional distribution, including quantile predictions and prediction intervals, provides a notable improvement in geospatial machine learning. The \textbf{PE-GQNN} provides a solid foundation for future advancements in the field of geospatial machine learning.

\vspace{.1cm}
\textbf{Limitations:}
\hspace{.15cm}
Although the Lipschitz Monotonic Networks (LMN) \citep{nolte2023} can approximate all monotonic Lipschitz bounded functions, many common quantile functions do not satisfy this condition. This often occurs because the derivative $\partial q(\tau) / \partial \tau$ tends to $\infty$ as $\tau$ approaches 1, and to $- \infty$ as $\tau$ approaches 0. The practical implications of this limitation when using LMN blocks remain to be fully understood. However, based on our experiments, it does not appear to pose a major concern. LMNs are unlikely to struggle when approximating truncated distributions, where the support is slightly restricted and the derivative $\partial q(\tau) / \partial \tau$ remains bounded. Naturally, any advances in monotonic neural network architectures would directly enhance the effectiveness of our approach.

\bibliography{Bibliography-V2.bib}

\begin{thebibliography}{}

\bibitem[Anselin, 1995]{anselin1995}
Anselin, L. (1995).
\newblock Local indicators of spatial association - {LISA}.
\newblock {\em Geographical analysis}, 27(2):93--115.

\bibitem[Anselin, 2022]{anselin2022}
Anselin, L. (2022).
\newblock Spatial econometrics.
\newblock {\em Handbook of Spatial Analysis in the Social Sciences}, pages 101--122.

\bibitem[Appleby et~al., 2020]{appleby2020}
Appleby, G., Liu, L., and Liu, L.-P. (2020).
\newblock Kriging convolutional networks.
\newblock In {\em Proceedings of the AAAI Conference on Artificial Intelligence}, volume 34(04), pages 3187--3194.

\bibitem[Bao et~al., 2024]{bao2024}
Bao, L.-L., Zhang, J., and Zhang, C. (2024).
\newblock Spatial multi-attention conditional neural processes.
\newblock {\em Neural networks : the official journal of the International Neural Network Society}, 173:106201.

\bibitem[Bi et~al., 2023]{bi2023}
Bi, K., Xie, L., Zhang, H., Chen, X., Gu, X., and Tian, Q. (2023).
\newblock Accurate medium-range global weather forecasting with 3d neural networks.
\newblock {\em Nature}, 619(7970):533--538.

\bibitem[Cressie et~al., 2022]{cressie+sz22}
Cressie, N., Sainsbury-Dale, M., and Zammit-Mangion, A. (2022).
\newblock Basis-function models in spatial statistics.
\newblock {\em Annual Review of Statistics and Its Application}, 9:373--400.

\bibitem[Cressie and Wikle, 2011]{cressie+w11}
Cressie, N. and Wikle, C. (2011).
\newblock {\em Statistics for Spatio-Temporal Data}.
\newblock CourseSmart Series. Wiley.

\bibitem[Datta et~al., 2016]{datta2016}
Datta, A., Banerjee, S., Finley, A.~O., and Gelfand, A.~E. (2016).
\newblock Hierarchical nearest-neighbor {Gaussian} process models for large geostatistical datasets.
\newblock {\em Journal of the American Statistical Association}, 111(514):800--812.

\bibitem[Derrow-Pinion et~al., 2021]{derrow2021}
Derrow-Pinion, A., She, J., Wong, D., Lange, O., Hester, T., Perez, L., Nunkesser, M., Lee, S., Guo, X., Wiltshire, B., et~al. (2021).
\newblock Eta prediction with graph neural networks in google maps.
\newblock In {\em Proceedings of the 30th ACM international conference on information \& knowledge management}, pages 3767--3776.

\bibitem[Fey and Lenssen, 2019]{pyg2019}
Fey, M. and Lenssen, J.~E. (2019).
\newblock Fast graph representation learning with pytorch geometric.

\bibitem[Furrer et~al., 2006]{furrer+gn06}
Furrer, R., Genton, M.~G., and Nychka, D. (2006).
\newblock Covariance tapering for interpolation of large spatial datasets.
\newblock {\em Journal of Computational and Graphical Statistics}, 15(3):502--523.

\bibitem[Hamilton et~al., 2017]{hamilton2017}
Hamilton, W., Ying, Z., and Leskovec, J. (2017).
\newblock Inductive representation learning on large graphs.
\newblock In Guyon, I., Luxburg, U.~V., Bengio, S., Wallach, H., Fergus, R., Vishwanathan, S., and Garnett, R., editors, {\em Advances in Neural Information Processing Systems}, volume~30. Curran Associates, Inc.

\bibitem[Hooker et~al., 2018]{hooker2018}
Hooker, J., Duveiller, G., and Cescatti, A. (2018).
\newblock A global dataset of air temperature derived from satellite remote sensing and weather stations.
\newblock {\em Scientific Data}, 5(1):1--11.

\bibitem[Kashyap et~al., 2022]{kashyap2022}
Kashyap, A.~A., Raviraj, S., Devarakonda, A., Nayak~K, S.~R., KV, S., and Bhat, S.~J. (2022).
\newblock Traffic flow prediction models--a review of deep learning techniques.
\newblock {\em Cogent Engineering}, 9(1):2010510.

\bibitem[Katzfuss and Guinness, 2021]{katzfuss+g21}
Katzfuss, M. and Guinness, J. (2021).
\newblock {A general framework for Vecchia approximations of Gaussian processes}.
\newblock {\em Statistical Science}, 36(1):124 -- 141.

\bibitem[Kaul et~al., 2013]{kaul2013}
Kaul, M., Yang, B., and Jensen, C.~S. (2013).
\newblock Building accurate 3d spatial networks to enable next generation intelligent transportation systems.
\newblock In {\em 2013 IEEE 14th International Conference on Mobile Data Management}, volume~1, pages 137--146. IEEE.

\bibitem[Kingma and Ba, 2015]{kingma2015}
Kingma, D. and Ba, J. (2015).
\newblock {ADAM: A} method for stochastic optimization.
\newblock In {\em International Conference on Learning Representations (ICLR)}, San Diega, CA, USA.

\bibitem[Kipf and Welling, 2017]{kipf2017}
Kipf, T.~N. and Welling, M. (2017).
\newblock Semi-supervised classification with graph convolutional networks.
\newblock In {\em International Conference on Learning Representations (ICLR)}.

\bibitem[Klemmer and Neill, 2021]{klemmer2021}
Klemmer, K. and Neill, D.~B. (2021).
\newblock Auxiliary-task learning for geographic data with autoregressive embeddings.
\newblock In {\em Proceedings of the 29th International Conference on Advances in Geographic Information Systems}, pages 141--144.

\bibitem[Klemmer et~al., 2023]{klemmer2023}
Klemmer, K., Safir, N.~S., and Neill, D.~B. (2023).
\newblock Positional encoder graph neural networks for geographic data.
\newblock In {\em International Conference on Artificial Intelligence and Statistics}, pages 1379--1389. PMLR.

\bibitem[Koenker and Bassett~Jr, 1978]{koenker1978}
Koenker, R. and Bassett~Jr, G. (1978).
\newblock Regression quantiles.
\newblock {\em Econometrica: journal of the Econometric Society}, pages 33--50.

\bibitem[Kuleshov and Deshpande, 2022]{kuleshov2022}
Kuleshov, V. and Deshpande, S. (2022).
\newblock Calibrated and sharp uncertainties in deep learning via density estimation.
\newblock In {\em ICML}, pages 11683--11693.

\bibitem[Kuleshov et~al., 2018]{kuleshov2018}
Kuleshov, V., Fenner, N., and Ermon, S. (2018).
\newblock Accurate uncertainties for deep learning using calibrated regression.
\newblock In {\em International conference on machine learning}, pages 2796--2804. PMLR.

\bibitem[Lindgren et~al., 2011]{lindgren+rl11}
Lindgren, F., Rue, H., and Lindstr{\"o}m, J. (2011).
\newblock {An explicit link between Gaussian fields and Gaussian Markov random fields: The stochastic partial differential equation approach}.
\newblock {\em Journal of the Royal Statistical Society Series B: Statistical Methodology}, 73(4):423--498.

\bibitem[Lv et~al., 2014]{lv2014}
Lv, Y., Duan, Y., Kang, W., Li, Z., and Wang, F.-Y. (2014).
\newblock Traffic flow prediction with big data: A deep learning approach.
\newblock {\em IEEE Transactions on Intelligent Transportation Systems}, 16(2):865--873.

\bibitem[Mai et~al., 2020]{mai2020}
Mai, G., Janowicz, K., Yan, B., Zhu, R., Cai, L., and Lao, N. (2020).
\newblock Multi-scale representation learning for spatial feature distributions using grid cells.
\newblock In {\em International Conference on Learning Representations}.

\bibitem[Nolte et~al., 2023]{nolte2023}
Nolte, N., Kitouni, O., and Williams, M. (2023).
\newblock Expressive monotonic neural networks.
\newblock In {\em The Eleventh International Conference on Learning Representations}.

\bibitem[Pace and Barry, 1997]{pace1997}
Pace, R.~K. and Barry, R. (1997).
\newblock Sparse spatial autoregressions.
\newblock {\em Statistics \& Probability Letters}, 33(3):291--297.

\bibitem[Paszke et~al., 2019]{pytorch2019}
Paszke, A., Gross, S., Massa, F., Lerer, A., Bradbury, J., Chanan, G., Killeen, T., Lin, Z., Gimelshein, N., Antiga, L., et~al. (2019).
\newblock Pytorch: An imperative style, high-performance deep learning library.
\newblock {\em Advances in Neural Information Processing Systems}, 32.

\bibitem[Rasmussen and Williams, 2006]{rasmussen+w06}
Rasmussen, C.~E. and Williams, C. K.~I. (2006).
\newblock {\em Gaussian Processes for Machine Learning}.
\newblock The MIT Press.

\bibitem[Rodrigues and Pereira, 2020]{rodrigues2020}
Rodrigues, F. and Pereira, F.~C. (2020).
\newblock Beyond expectation: Deep joint mean and quantile regression for spatiotemporal problems.
\newblock {\em IEEE Transactions on Neural Networks and Learning Systems}, 31(12):5377--5389.

\bibitem[Si et~al., 2022]{si2022}
Si, P., Kuleshov, V., and Bishop, A. (2022).
\newblock Autoregressive quantile flows for predictive uncertainty estimation.
\newblock In {\em International Conference on Learning Representations}.

\bibitem[Sreenivasa and Nirmala, 2019]{sreenivasa2019}
Sreenivasa, B. and Nirmala, C. (2019).
\newblock Hybrid location-centric e-commerce recommendation model using dynamic behavioral traits of customer.
\newblock {\em Iran Journal of Computer Science}, 2(3):179--188.

\bibitem[Tagasovska and Lopez-Paz, 2019]{Tagasovska2019}
Tagasovska, N. and Lopez-Paz, D. (2019).
\newblock {\em Single-model uncertainties for deep learning}.
\newblock Curran Associates Inc., Red Hook, NY, USA.

\bibitem[Vaswani et~al., 2017]{vaswani2017}
Vaswani, A., Shazeer, N., Parmar, N., Uszkoreit, J., Jones, L., Gomez, A.~N., Kaiser, {\L}., and Polosukhin, I. (2017).
\newblock Attention is all you need.
\newblock {\em Advances in Neural Information Processing Systems}, 30.

\bibitem[Vecchia, 1998]{vecchia98}
Vecchia, A.~V. (1998).
\newblock {Estimation and model identification for continuous spatial processes}.
\newblock {\em Journal of the Royal Statistical Society: Series B (Methodological)}, 50(2):297--312.

\bibitem[Veli{\v{c}}kovi{\'{c}} et~al., 2018]{velickovic2018}
Veli{\v{c}}kovi{\'{c}}, P., Cucurull, G., Casanova, A., Romero, A., Li{\`{o}}, P., and Bengio, Y. (2018).
\newblock {Graph attention networks}.
\newblock {\em International Conference on Learning Representations}.

\bibitem[Wu et~al., 2022]{wu2022}
Wu, L., Cui, P., Pei, J., Zhao, L., and Guo, X. (2022).
\newblock Graph neural networks: {Foundation}, frontiers and applications.
\newblock In {\em Proceedings of the 28th ACM SIGKDD Conference on Knowledge Discovery and Data Mining}, pages 4840--4841.

\bibitem[Xu et~al., 2020]{xu2020}
Xu, S., Fu, X., Cao, J., Liu, B., and Wang, Z. (2020).
\newblock Survey on user location prediction based on geo-social networking data.
\newblock {\em World Wide Web}, 23(3):1621--1664.

\end{thebibliography}
\bibliographystyle{apalike}

\appendix
\section{Additional technical details}

\subsection{Quantile Neural Networks and recalibration} 
\label{sub:appendix}

\begin{figure}[h]
     \centering
     \begin{subfigure}[b]{0.47\textwidth}
         \centering
         \includegraphics[width=\textwidth]{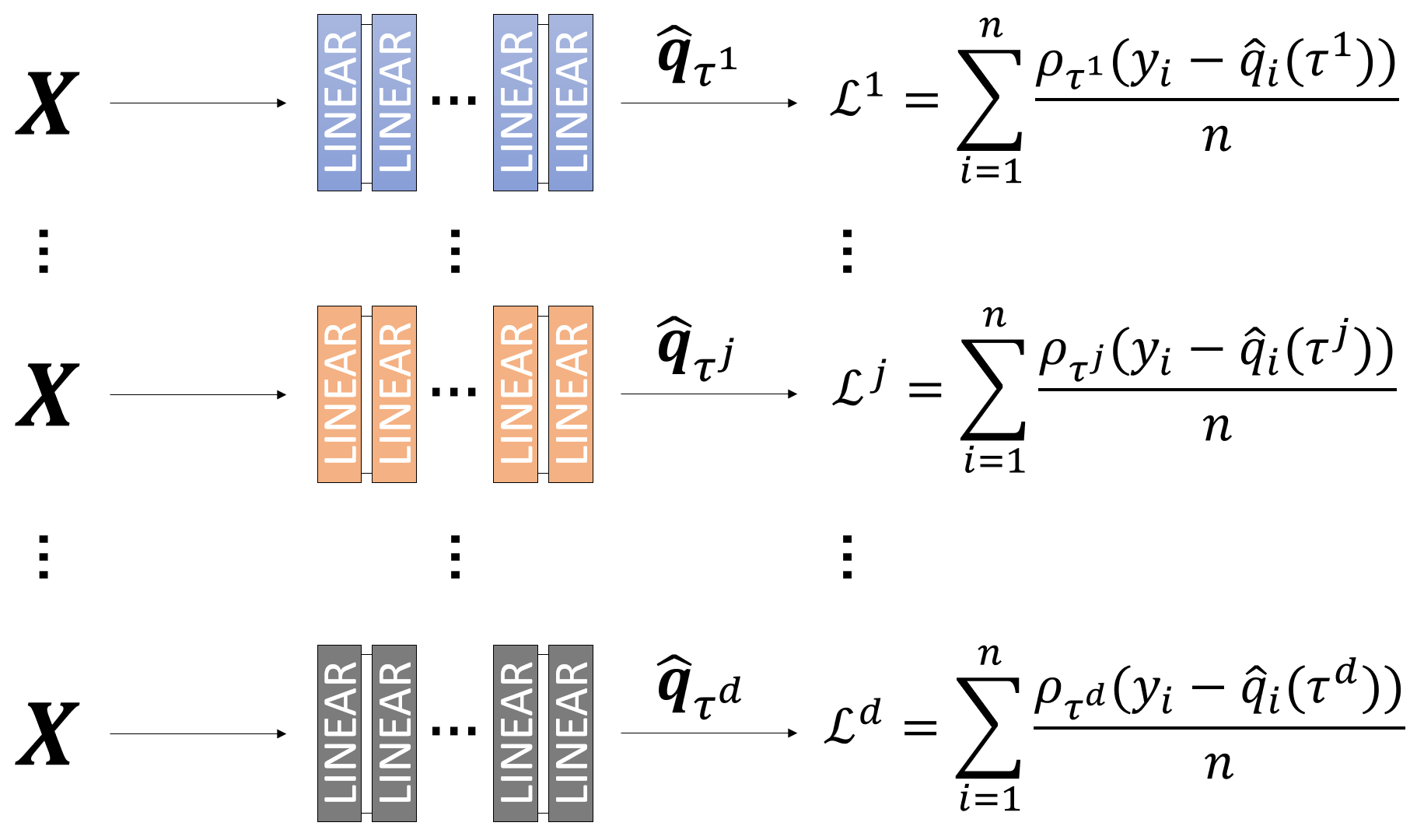}
         \caption{Non-linear quantile regression using NN.}
         \label{fig:nonlinear}
     \end{subfigure}
     \hfill
     \begin{subfigure}[b]{0.47\textwidth}
         \centering
         \includegraphics[width=\textwidth]{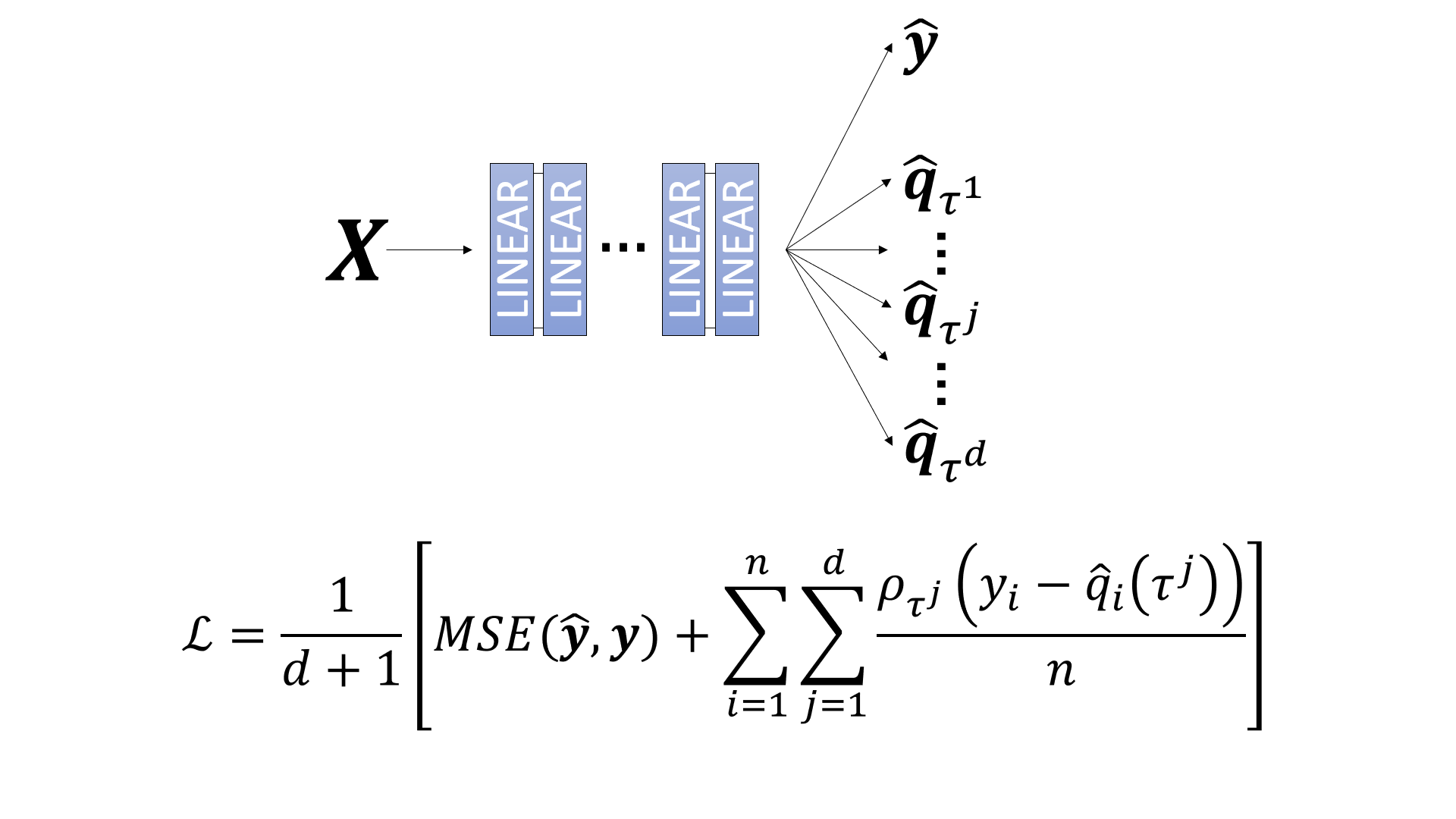}
         \caption{Non-linear multiple quantile regression.}
         \label{fig:nonlinear-multi}
     \end{subfigure}
     \hfill
     \vspace{1cm}
     \begin{subfigure}[b]{0.47\textwidth}
         \centering
         \includegraphics[width=\textwidth]{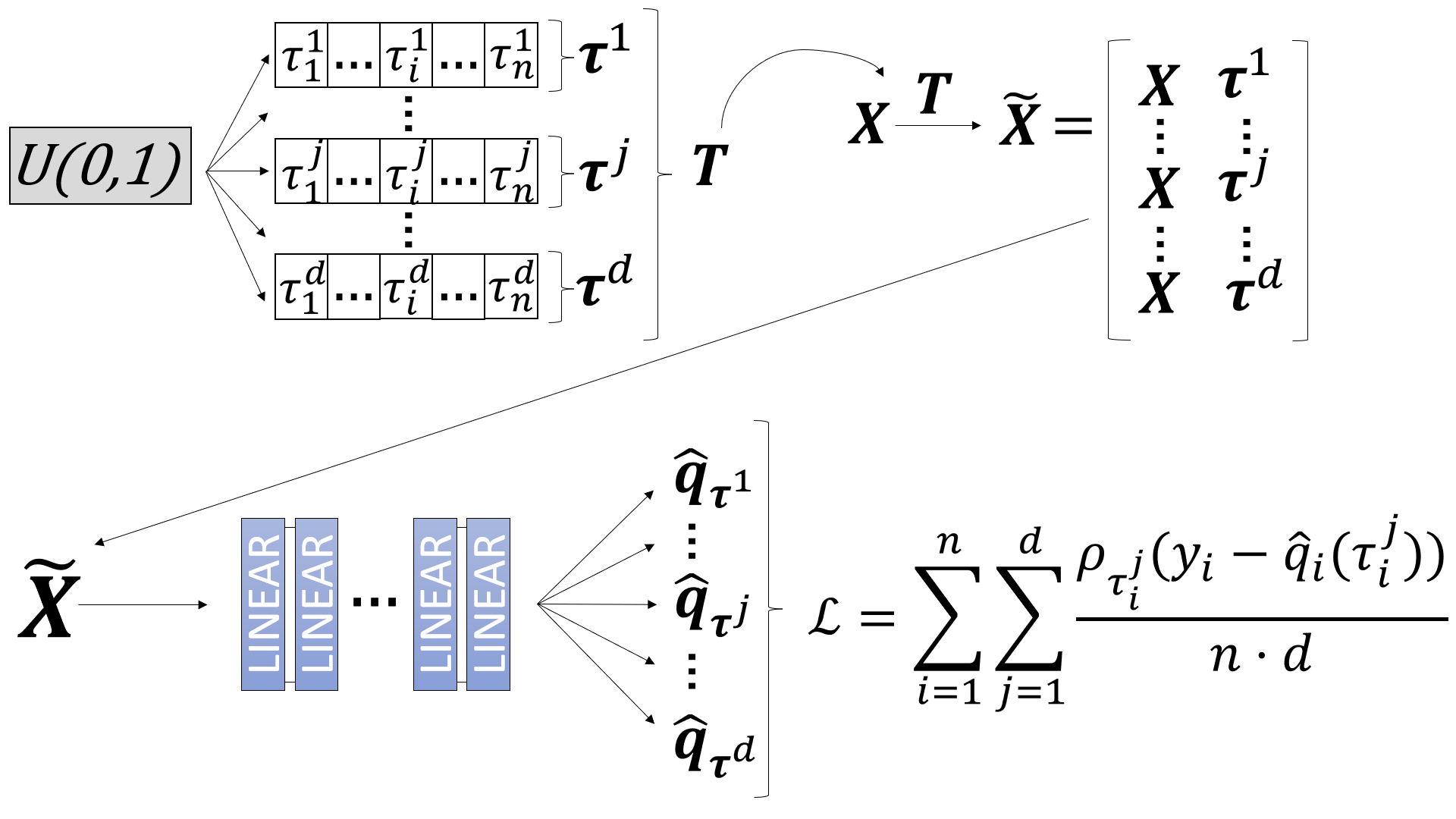}
         \caption{Non-linear quantile function regression.}
         \label{fig:si}
     \end{subfigure}
     \hfill
     \begin{subfigure}[b]{0.47\textwidth}
         \centering
         \includegraphics[width=\textwidth]{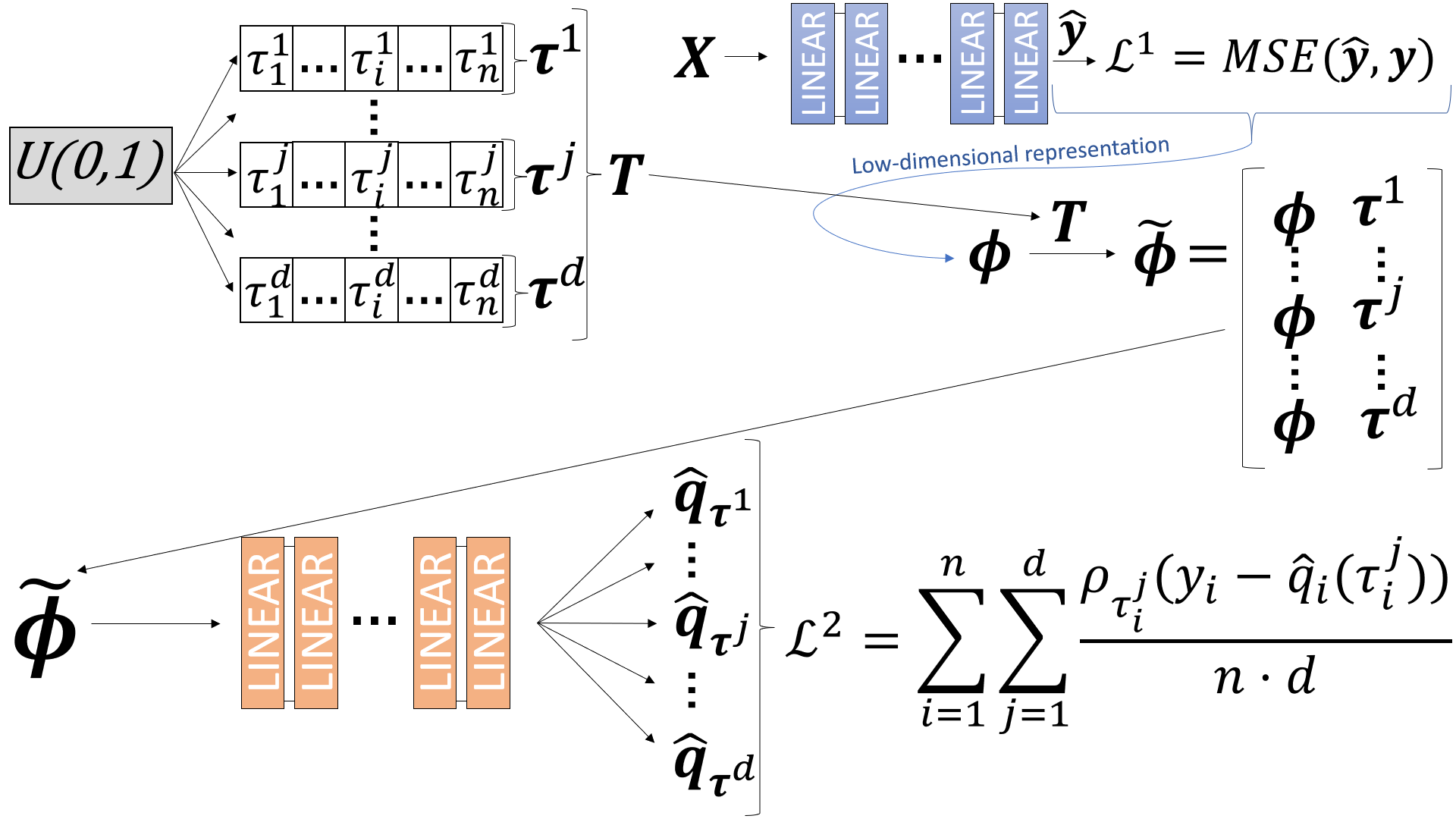}
         \caption{Two-step density estimation.}
         \label{fig:kuleshov}
     \end{subfigure}
        \caption{(a) For each quantile of interest, a separate NN is trained. (b) 
        \citet{rodrigues2020}: one NN outputs $d+1$ predictions: one for the expectation and $d$ for the quantiles. (c) \citet{si2022}: a single NN trained to predict \textit{any} generic quantile of the conditional distribution. (d) \citet{kuleshov2022}: two-step procedure: the first model outputs a low-dimensional representation of the conditional distribution, which a recalibrator then uses to produce calibrated predictions.}
        \label{fig:quantile-panel}
\end{figure}

Figure \ref{fig:quantile-panel} provides a visual overview of some of the most closely related approaches.

\vspace{.1cm}
\textbf{Number of Monte Carlo samples:}
\hspace{.15cm}
When applying the framework proposed by \citet{si2022}, we chose to use $d=1$ for the $\tau$ values. Let $\mathcal{L}(\theta, \tau, \bm{x}, y)$ be the loss function for a given quantile $\tau \sim \text{U}(0, 1)$ and an observed pair $(\bm{x}, y) \sim D_{\text{data}}$, where $D_{\text{data}}$ denotes the full data generative process. On each training iteration, we minimize $\mathcal{L}_B$, which, by the Law of Large Numbers, converges to $\tilde{\mathcal{L}}(\theta) = \mathbb{E}_{\tau, \bm{x}, y} \mathcal{L}(\theta, \tau, \bm{x}, y)$, as the batch size, ${n_B}$, goes to $\infty$. Therefore, the gradients converge to the same value for any $d$, provided that ${n_B} \to \infty$. This choice $(d=1)$, which is also used by \citet{Tagasovska2019}, simplifies the implementation without sacrificing performance, as shown in Section \ref{sec:experiments}.

\subsection{Positional Encoder}
\label{sub:PE}

Inspired by the Transformer architecture \citep{vaswani2017}, \citet{mai2020} introduced the Positional Encoder (PE) for geographic data. 
The PE maps the geographic coordinate vector $\bm{c} = (c_{1}, c_{2}) \in \mathbb{R}^2$ of a single datapoint, typically representing latitude and longitude, to a high-dimensional embedding using a set of deterministic sinusoidal transformations followed by a fully connected neural network. 

The encoded spatial embedding $\bm{c}^{\text{emb}} \in \mathbb{R}^d$ is computed as:

\begin{equation}
    \bm{c}^{\text{emb}} = PE(\bm{c}, \sigma_{\min}, \sigma_{\max}, \Theta_{\text{PE}}) = NN(ST(\bm{c}, \sigma_{\min}, \sigma_{\max}), \Theta_{\text{PE}}),
    \label{eq:pe_nn_point}
\end{equation}

where $NN(\cdot, \Theta_{\text{PE}})$ denotes a fully connected neural network with parameters $\Theta_{\text{PE}}$, and $ST(\cdot)$ is a sinusoidal transformation defined by:

\begin{equation}
    ST(\bm{c}, \sigma_{\min}, \sigma_{\max}) = \left(ST_1(\bm{c}, \sigma_{\min}, \sigma_{\max}) \\;  \ldots \\; ST_S(\bm{c}, \sigma_{\min}, \sigma_{\max}) \right),
\end{equation}

with $\sigma_{\min}$ and $\sigma_{\max}$ as hyperparameters. Each component $ ST_j(\bm{c}, \sigma_{\min}, \sigma_{\max})$, for $j \in \{1, \ldots, S\}$, is given by:

\begin{equation}
     ST_j(\bm{c}, \sigma_{\min}, \sigma_{\max}) = \left( 
        \sin\left( \frac{2\pi c_1}{\sigma_j} \right), 
        \cos\left( \frac{2\pi c_1}{\sigma_j} \right),
        \sin\left( \frac{2\pi c_2}{\sigma_j} \right), 
        \cos\left( \frac{2\pi c_2}{\sigma_j} \right)
    \right),
    \label{eq:pe_st_point}
\end{equation}

where the values $\sigma_1, \ldots, \sigma_S$ form a logarithmically spaced grid between $\sigma_{\min}$ and $\sigma_{\max}$, computed as:

\begin{equation}
    \sigma_j = \sigma_{\min} \cdot \left( \frac{\sigma_{\max}}{\sigma_{\min}} \right)^{\frac{j - 1}{S - 1}}.
    \label{eq:sigma_j}
\end{equation}

The result of $ST(\bm{c}, \cdot)$ is a $4S$-dimensional vector that encodes spatial patterns at multiple scales. This vector is then passed through the neural network $NN$ to produce the final embedding $\bm{c}^{\text{emb}}$.

\newpage
\subsection{Spatial visualization of the California Housing dataset predictions}

\begin{figure}[h]
     \centering
     \begin{subfigure}[b]{0.47\textwidth}
         \centering
         \includegraphics[width=\textwidth]{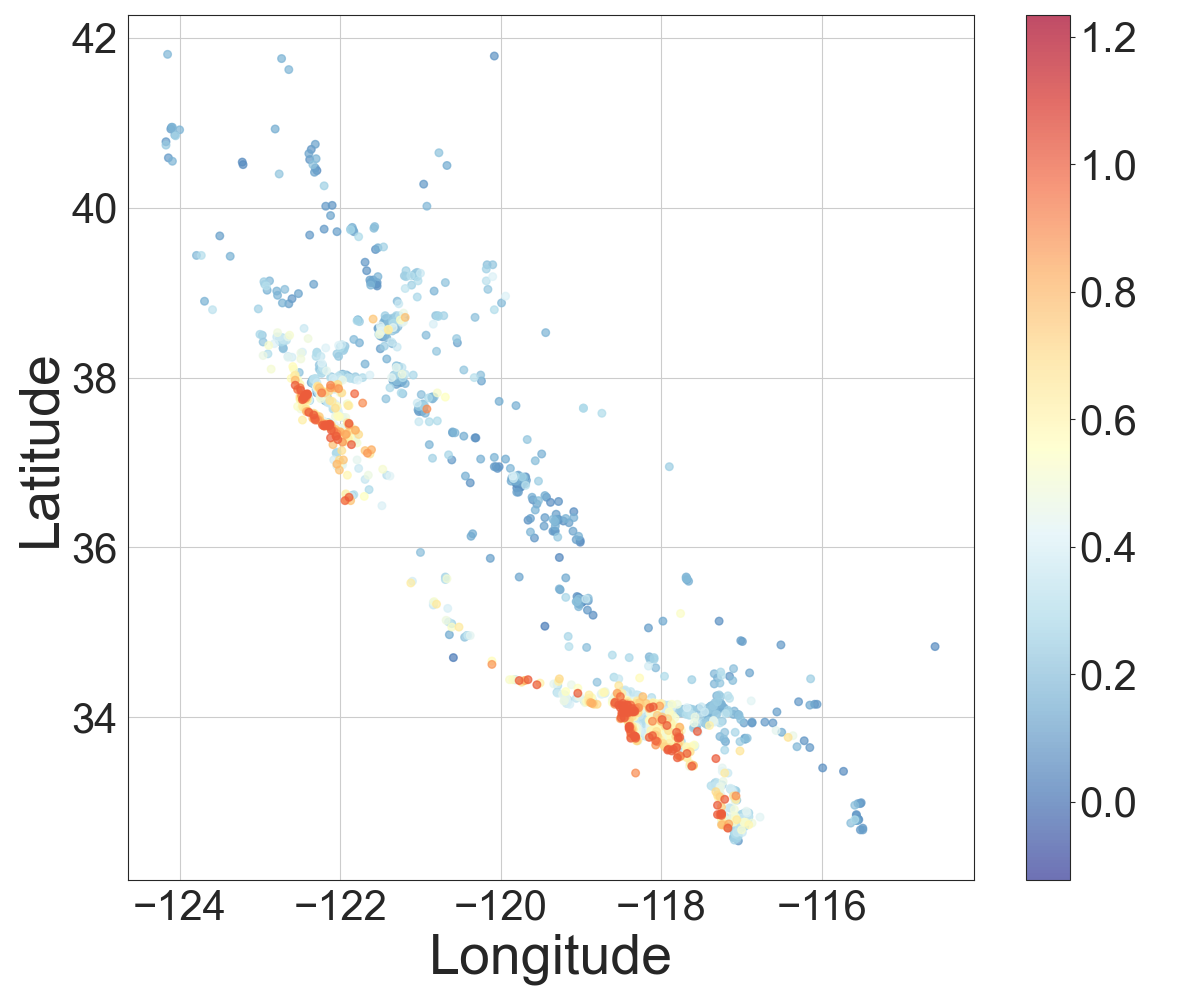}
         \caption{Test values.}
         \label{fig:target}
     \end{subfigure}
     \hfill
     \begin{subfigure}[b]{0.47\textwidth}
         \centering
         \includegraphics[width=\textwidth]{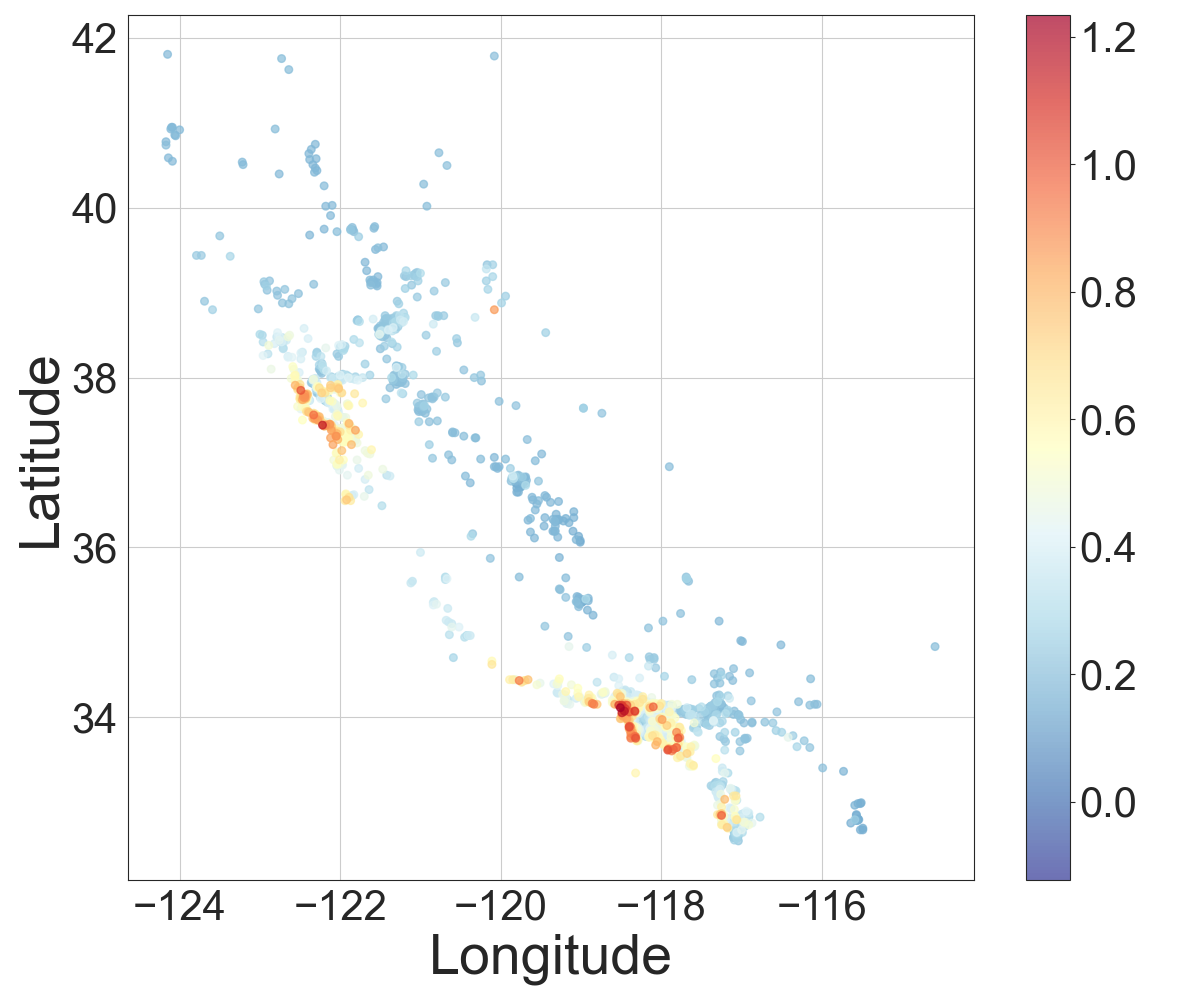}
         \caption{PE-GSAGE predictions.}
         \label{fig:PE-GSAGE}
     \end{subfigure}
     \vspace{1cm}
     \begin{subfigure}[b]{0.47\textwidth}
         \centering
         \includegraphics[width=\textwidth]{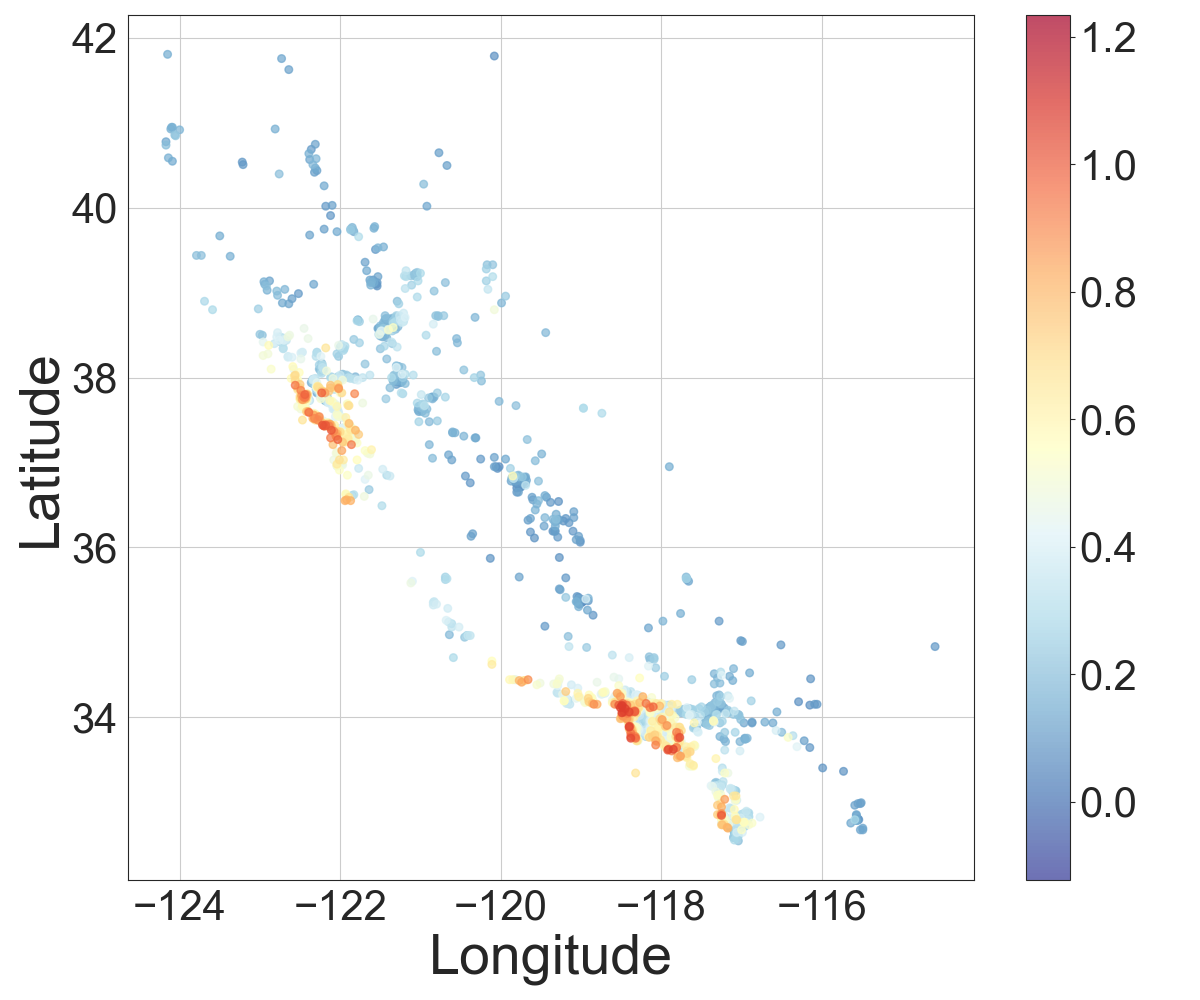}
         \caption{PE-GQSAGE predictions.}
         \label{fig:_pegqsage}
     \end{subfigure}
     \hfill
     \begin{subfigure}[b]{0.47\textwidth}
         \centering
         \includegraphics[width=\textwidth]{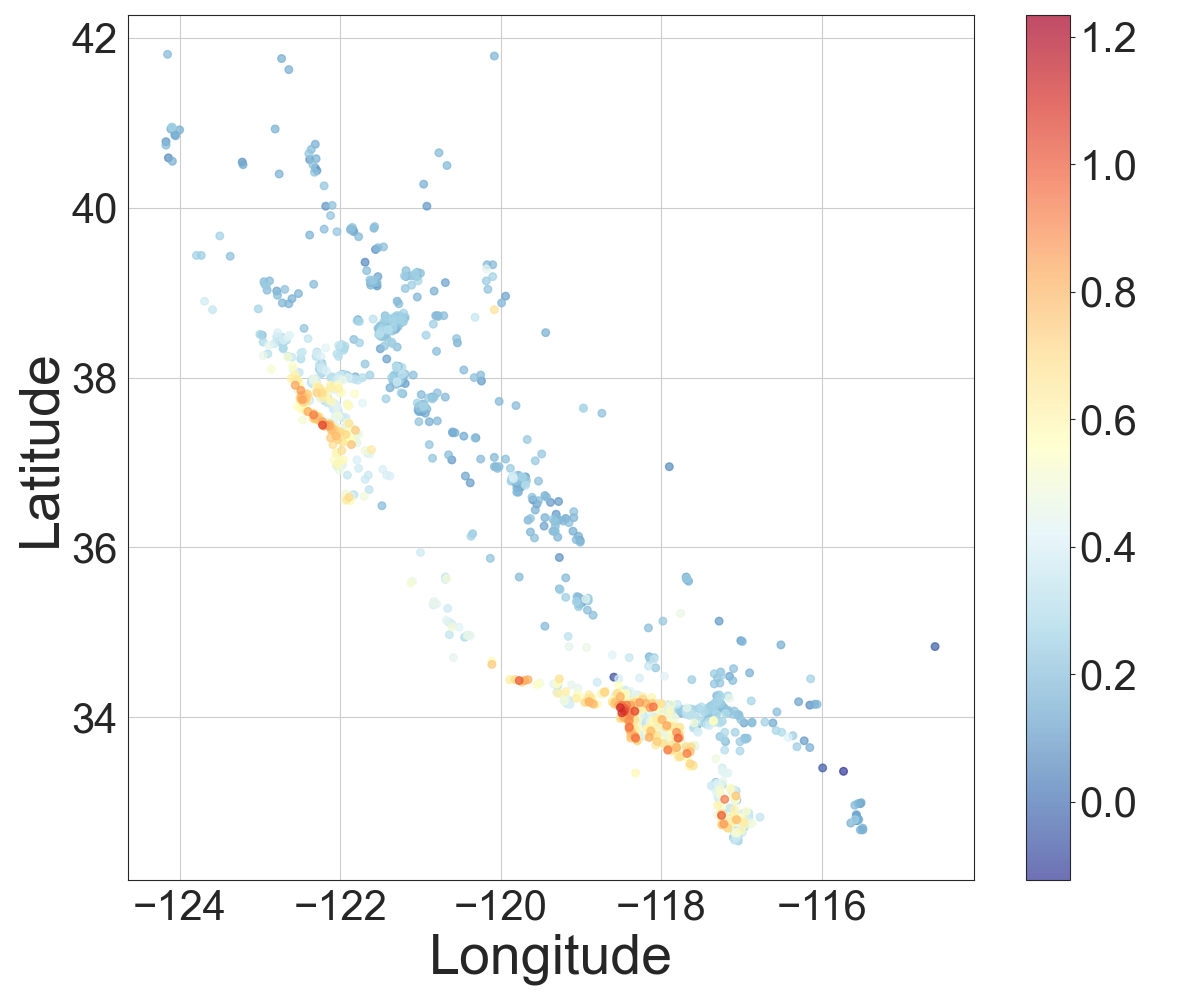}
         \caption{SMACNP predictions.}
         \label{fig:smacnp}
     \end{subfigure}
        \caption{Geographical maps of the predicted results on the California Housing test dataset.}
        \label{fig:maps}
\end{figure}

Maps of the test values and predictions of selected methods are illustrated in Figure \ref{fig:maps}. Visually, the predictions provided by \textbf{PE-GQNN} appear to be closest to the actual data behavior, particularly in the major cities.

\newpage
\subsection{Architectural details}
\label{sub:details}

To facilitate a clear and detailed comparison between \textbf{PE-GQNN} and PE-GNN, we present the architectures for the California Housing dataset in Tables \ref{tab:pegqsage_architecture} and \ref{tab:pegsage_architecture}. The same architectures are used for the Air Temperature dataset, differing only in the number of node features ($p = 1$). For the 3D Road dataset, which is an interpolation task without node features, the GNN block is omitted.

\begin{table}[ht]
\centering
\resizebox{.93\textwidth}{!}{
\begin{tabular}{l|r|l|l}
\hline
\textbf{Layer (type \& shape)} & \textbf{\# Param.} & \textbf{Notes} & \textbf{Activation}\\
\hline
\textbf{GNN block}: 6 → 32 & -- & -- &  -- \\
\quad Input: node features (dim = 6) & -- & Node features input & -- \\
\quad SAGEConv (hidden): 6 → 32 & -- & GraphConv layer & -- \\
\quad \quad Aggregation & 224 & GSAGE internal layer & -- \\
\quad \quad Update & 192 & GSAGE internal layer & ReLU \\
\quad SAGEConv (out): 32 → 32 & -- & GraphConv layer & -- \\
\quad \quad Aggregation & 1,056 & GSAGE internal layer & -- \\
\quad \quad Update & 1,024 & GSAGE internal layer & ReLU \\
\hline
\textbf{Positional encoder (PE) block}: 2 → 64 &  -- &  -- &  --\\
 \quad Input: coordinates (dim = 2) & -- & Spatial coordinates input & --  \\
 \quad Sinusoidal Transformation (ST): 2 → 64 & 0 & Fourier feature mapping & -- \\
 \quad Dropout regularization & -- & Dropout rate: p = 0.5 & -- \\
 \quad Linear (hidden): 64 → 128 & 8,320 & Feedforward layer& ReLU \\
 \quad Linear (hidden): 128 → 64 & 8,256 & Feedforward layer & Tanh \\
 \quad Linear (hidden): 64 → 32 & 2,080 & Feedforward layer & Tanh\\
 \quad Linear (out): 32 → 64 & 2,112 & Feedforward layer& Identity\\
 \hline
 \textbf{Concatenation}: GNN (32) + PE (64) & 0 & Combined vectors & --\\
\hline
\textbf{Fully-connected (FC) block}: 96 → 8 & -- & Dimension reduction block & --\\
\quad Linear (hidden): 96 → 32 & 3,104 & Feedforward layer & Tanh\\
\quad Linear (hidden): 32 → 16 & 528 & Feedforward layer & Tanh\\
\quad Linear (out): 16 → 8 & 136 & Feedforward layer & Identity\\
\hline
\textbf{Concatenation}: FC (8) + $\bar{y}$ (1) + $\tau$ (1) & 0 & Combined vectors & --\\
\hline
\textbf{Monotonic Layers}: 10 → 1 & -- & Final quantile regressor & -- \\
\quad Lipschitz Linear (hidden): 10 → 32 & 352 & Monotonic layer 1 & GroupSort(2) \\
\quad Lipschitz Linear (out): 32 → 1 & 33 & Monotonic layer 2 & Identity\\
\hline
\textbf{Total} & 27,417 & All trainable parameters & --\\
\hline
\end{tabular}
}
\caption{Architecture of PE-GQSAGE applied to the California Housing dataset, illustrating input/output dimensions, parameter counts, layer-level annotations, and activation functions.}
\label{tab:pegqsage_architecture}
\end{table}

\begin{table}[ht]
\centering
\resizebox{.93\textwidth}{!}{
\begin{tabular}{llll}
\hline
\textbf{Layer (type \& shape)} & \textbf{\# Param.} & \textbf{Notes} & \textbf{Activation} \\
\hline
\textbf{Positional Encoder (PE) block}: 2 → 64 &  -- & -- & -- \\
\quad Input: coordinates (dim = 2) & -- & Spatial coordinates input & -- \\
\quad Sinusoidal Transformation (ST): 2 → 64 & 0 & Fourier feature mapping & -- \\
\quad Dropout regularization & -- & Dropout rate: p = 0.5 & -- \\
\quad Linear (hidden): 64 → 128 & 8,320 & Feedforward layer & ReLU \\
\quad Linear (hidden): 128 → 64 & 8,256 & Feedforward layer & Tanh \\
\quad Linear (hidden): 64 → 32 & 2,080 & Feedforward layer & Tanh \\
\quad Linear (out): 32 → 64 & 2,112 & Feedforward layer & Identity \\
\hline
\textbf{Concatenation}: PE (64) + Features (6) & 0 & Combined input vector & -- \\
\hline
\textbf{GNN block}: 70 → 32 & -- & -- & -- \\
\quad Input: node features (dim = 70) & -- & PE + Node attributes & -- \\
\quad SAGEConv (hidden): 70 → 32 & -- & GraphConv layer & -- \\
\quad\quad Aggregation Linear & 2,272 & GSAGE internal layer & -- \\
\quad\quad Update Linear & 2,240 & GSAGE internal layer & ReLU \\
\quad SAGEConv (out): 32 → 32 & -- & GraphConv layer & -- \\
\quad\quad Aggregation Linear & 1,056 & GSAGE internal layer & -- \\
\quad\quad Update Linear & 1,024 & GSAGE internal layer & ReLU \\
\hline
\textbf{Fully-connected layers:} 32 → 1 & -- & -- & -- \\
\quad Linear (to target): 32 → 1 & 33 & Final output layer & Identity \\
\quad Linear (to Moran's I): 32 → 1 & 33 & Auxiliary output layer & Identity \\
\hline
\textbf{Total} & 27,426 & All trainable parameters & -- \\
\hline
\end{tabular}
}
\caption{Architecture of PE-GSAGE. Positional encodings derived from spatial coordinates are concatenated with node features prior to the GNN layers. This architecture follows the specification introduced by \citet{klemmer2023}.}
\label{tab:pegsage_architecture}
\end{table}

\end{document}